\documentclass[aps,rmp,reprint,amsmath,amssymb,graphicx,longbibliography]{revtex4-1}
\usepackage{tikz}
\usepackage{amsmath}
\usepackage{mathtools}
\usepackage{caption}
\usepackage{subcaption}
\usepackage[margin=1in]{geometry}
\usepackage{xcolor}
\usepackage{graphicx}
\graphicspath{{./}{figures/}}

\DeclareMathOperator{\LoG}{LoG}
\DeclareMathOperator{\DoG}{DoG}

\usepackage{bm}

\begin{document}

%\title{Towards a unified mathematical framework for information space}
\title{A Unified Framework for the Mechanics of Information \\ in Convolutional Neural Network Image Space}

\author{Aryan Shukla}
\author{Matthew Toews} 
\affiliation{Department of Systems Engineering, École de technologie supérieure, 1100
R. Notre Dame O., Montréal, QC H3C 1K3, Canada, matt.toews@gmail.com}

\date{\today{}}

\begin{abstract}
This paper introduces a unified mathematical framework for modeling information propagation through convolutional neural networks (CNNs), with the aim of connecting descriptions of physical space and information space.

A correspondence is presented linking discrete filter symmetry and the relativistic energy--momentum relation under the widely used nonlinear rectified convolution operation. Specifically, symmetric filter components (e.g. the sum $\Sigma = [1,1]$)  operate analogously to rest energy $mc^2$ in preserving the image centre of mass (e.g. isotropic diffusion), whereas antisymmetric components (e.g. the gradient $\nabla = [-1,1]$) operate analogously to the momentum term $pc$ in generally inducing a displacement (e.g. vibration or translation). For typical small discrete filters, this displacement is determined by the ratio of antisymmetric to total filter energy, analogously to how the displacement of a relativistic particle relates to a Lorentz transform with beta parameter $\beta = \frac{v}{c}=\frac{pc}{E}$ equal to the ratio of momentum $pc$ to total energy $E$.

Repeated filtering leads to the Gaussian scale-space and emergent scale-invariant features. These constructions share a Laplacian-driven structure with the classical heat (diffusion) equation and, via standard mathematical correspondences, with the Schr\"odinger equation and aspects of the Friedmann equations, together with emergent Morse topological structure. Demonstrations in 3D images reveal blob-like, scale-invariant Morse critical points in images spanning a wide range of physical scales, including organic sugar molecules and inorganic silicon crystals, human and primate brains in magnetic resonance images (MRI), galaxies and the cosmic microwave background (CMB).
\end{abstract}

\maketitle

\tableofcontents{}

\section{Introduction}
\label{intro}

Symmetry has been used in physics to derive unifying models, and here we use symmetry to present a unified theory of the mechanics of information in artificial CNNs under the widely used rectified convolution operation. From physics we draw inspiration from Witten~\cite{witten1982supersymmetry}, who proposed supersymmetry driven by symmetric (bosonic) and antisymmetric (fermionic) operators leading to emergent Morse topology, and from Winterberg's minimal particle model~\cite{winterberg1994planck,winterberg2003planck}---a non-mainstream but conceptually related proposal driven by finite sum and difference operators with symmetric and antisymmetric modes of propagation. From image processing we draw inspiration from the CNN~\cite{lecun1989handwritten}, which led to modern AI with highly parallelized GPU training~\cite{krizhevsky2012imagenet}, and from binomial filtering~\cite{wells1986efficient}, which leads to the Gaussian scale-space~\cite{koenderink1984structure} and emergent Morse topology~\cite{damon1995local}, including scale-invariant image keypoints~\cite{lowe2004distinctive}.

Our theory begins with primary sum $[1,1]$ and difference $[-1,1]$ operators in discrete real-valued 1D image space, which may be generalized to 2D and 3D space via symmetry and to larger sizes via superposition, here via the discrete cosine transform (DCT). The primary novelty of our work is in studying the nonlinear rectification operation, introduced relatively recently~\cite{nair2010rectified} and now ubiquitous in neural network filtering, which as we show can endow information with an analogue of relativistic momentum in image space.

Our work follows an analogy between the effect of CNN filtering on information and state evolution in quantum mechanics. Just as a quantum state evolves under linear operators and, in relativistic settings, is constrained by the maximum speed of light $c$, the CNN activation image evolves under linear filtering operators bounded by a maximum displacement per layer set by the discrete filter width. However, while quantum mechanics typically involves complex Hermitian operators and matrix multiplication, CNN filtering involves real-valued discrete operators and convolution. Furthermore, in CNN filtering, convolution is followed by nonlinear rectification (the rectified linear unit),
\[
\mathrm{ReLU}(x) \;=\; \max(0,\,x).
\]
Rectification enforces non-negativity and restricts image information to the positive orthant of image space---a nonlinear constraint that, by loose analogy, plays a role similar to forbidding kinematically disallowed regions in relativistic energy--momentum space (the mass shell), though the geometries differ. The novelty of our work is in analyzing this nonlinear system in terms of relativistic energy--momentum theory, demonstrating Lorentz-like displacement behaviour and topological structure emerging from CNN filtering. The study of information mechanics may thus help clarify structure in information space in a spirit similar to how quantum mechanics clarified structure in physical space.

\section{Propagation in 1D Image Space}

We begin by introducing our theory in the minimal context of 1D scalar image space and $2$-pixel filter size to provide intuition with respect to well-known binomial filtering. We then generalize it to 2D and 3D image space and larger filter sizes using symmetry in the following sections. 

Consider a simple 1D image with a single non-zero pixel impulse:
\[
I^{(0)} \;=\; \begin{bmatrix}0 & 0 & 1 & 0 & 0\end{bmatrix}
\]

In this minimal context, information propagates within pixel space under the application of two hypothesized operators $\Sigma = [1,1]$ and $\nabla = [-1,1]$ spanning the space of real-valued 2-pixel filters in 1D space. Here we observe how these result in two modes of information propagation under rectified convolution, which may be described as diffusion and vibration, respectively.

\bigskip
\noindent\textbf{1. Diffusion mode (\(\Sigma\) operator)}  

Define the finite integration or sum operator
\[
\Sigma = \begin{bmatrix}1 & 1\end{bmatrix}
\]
Sequential convolution of an impulse at each iteration \(t\) with $\Sigma$
\[
I^{(t)} = I^{(t-1)} * \Sigma
\]
results in the well-known binomial pyramid:
\begin{gather*}
I^{(0)} = \begin{bmatrix}0 & 0 & \textcolor{red}{\textbf{1}} & 0 & 0\end{bmatrix},\\
I^{(1)} = \begin{bmatrix}0 & 0 & \textcolor{red}{\textbf{1}} & \textcolor{red}{\textbf{1}} & 0 & 0\end{bmatrix},\\
I^{(2)} = \begin{bmatrix}0 & 0 & \textcolor{red}{\textbf{1}} & 2 & \textcolor{red}{\textbf{1}} & 0 & 0\end{bmatrix},\\
I^{(3)} = \begin{bmatrix}0 & 0 & \textcolor{red}{\textbf{1}} & 3 & 3 & \textcolor{red}{\textbf{1}} & 0 & 0\end{bmatrix},\\
I^{(4)} = \begin{bmatrix}0 & 0 & \textcolor{red}{\textbf{1}} & 4 & 6 & 4 & \textcolor{red}{\textbf{1}} & 0 & 0\end{bmatrix}, \\ \dots
\end{gather*}

\begin{figure}[h]
    \centering
    \includegraphics[width=0.5\textwidth]{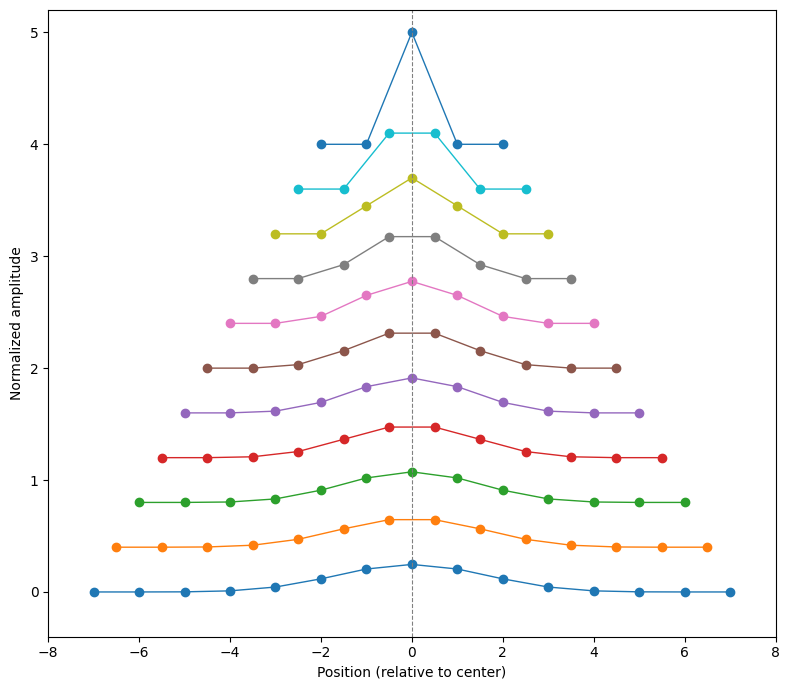}
    \caption{Stacked Distributions from Successive Convolutions.}
    \label{fig:diffusion}
\end{figure}

As can be seen from the evolution of the image, $I^{(0)}$, information spreads symmetrically in the form of the well-known binomial pyramid, approximating a discrete diffusion process. As shown in Figure~\ref {fig:diffusion}, when appropriately normalized, the binomial pyramid approaches a Gaussian density with variance $\sigma^2 \propto t$ as the number of iterations approaches infinity $t \rightarrow \infty$. Note that rectification does not affect the binomial pyramid constructed from non-negative information and filters.

\bigskip
\noindent\textbf{2. Vibration mode (\(\nabla_{\pm}\) operators)}  

%(we are not using $x, y$ subscripts and instead we are using $+, -$ subscripts because $x,y$ are derivatives in perpendicular $x$ and $y$ directions, which can be only defined for a 2D image. Here we are dealing with the 1D case.

Define the finite difference or gradient operators
\[
\nabla_{+} = \begin{bmatrix}-1 & 1\end{bmatrix},
\qquad
\nabla_{-} = \begin{bmatrix}1 & -1\end{bmatrix}
\]
%where
%\[
%\nabla_{+} = -\nabla_{-}
%\]

% \subsection*{Probabilistic Vibration Mode as a Random Walk}
Rectified convolution of an impulse with a difference operator results in a shift left or right depending on the operator sign, this represents the primary departure from standard linear signal processing. A sequence of rectified convolutions thus results in a path of left or right steps defined by the gradient operator (e.g. either \(\nabla_{+}\) or \(\nabla_{-}\)) at each iteration. For example in a randomly initialized network, the path may be modelled as randomly selecting a left or right operator at each layer with probability $\alpha$ or $1-\alpha$, respectively. Assuming the selection to be independent and identically distributed (IID) over layers simplifies analysis and reflects a “coin‐flip” decision in a network with no systematic bias toward a particular gradient orientation.  In practice one may assume \(\alpha=0.5\) for an unbiased model or estimate \(\alpha\) from the empirical frequency of \(\nabla_{+}\) vs.\ \(\nabla_{-}\) in a trained network.

Let the total number of trials be $N=L+R$, the sum of left $L$ and right $R$ shifts. The net shift in a single direction (i.e.\ left) is thus $L-R$, which is bounded by $\pm N$:
\[
-N \le L-R \le N.
\]
If each layer independently selects a left shift with probability $\alpha$, then $L\sim\mathrm{Binomial}(N,\alpha)$ and $R=N-L$, so the net displacement is the transformed binomial random variable $L-R=2L-N$. The probability of a combination of left $L$ and right $R$ shifts is
\[
P(L,R) = \binom{L+R}{L} \alpha^{L} (1-\alpha)^{R}.
\]

Given equiprobable orientations (i.e.\ $\alpha = 0.5$), the most probable path contains equal numbers of left and right shifts $L = R = \frac{N}{2}$, i.e.\ a pure \textbf{``vibration''} mode where the impulse oscillates around the origin with zero net displacement. This corresponds to probability
\[
P\!\left(\frac{N}{2},\frac{N}{2}\right)
= \binom{N}{\frac{N}{2}}\,\alpha^{\frac{N}{2}}\,(1-\alpha)^{\frac{N}{2}}.
\]
% which still decays sub-exponentially with \(k\), unless \(\alpha = 0.5\).

To demonstrate, alternately convolving the impulse with \(\nabla_{+}\) and \(\nabla_{-}\) followed by ReLU
\[
I^{(t)} = \max\bigl(0,\;I^{(t-1)} * \nabla_{\pm}\bigr)
\]
leads to a path whose instantaneous steps reverse orientation; with equal numbers of each orientation the net displacement vanishes (pure vibration), whereas a sustained preference for one orientation yields net drift. A concrete alternating sequence is:
\begin{gather*}
I^{(0)} = \begin{bmatrix}0 & 0 & \textcolor{red}{\textbf{1}} & 0 & 0 \end{bmatrix},\\
I^{(1)} = \begin{bmatrix}0 & 0 &  \textcolor{red}{\textbf{1}} & 0 & 0 & 0\end{bmatrix},\\
I^{(2)} = \begin{bmatrix}0 & 0 & 0 &  \textcolor{red}{\textbf{1}} & 0 & 0 & 0\end{bmatrix},\\
I^{(3)} = \begin{bmatrix}0 & 0 & 0 & \textcolor{red}{\textbf{1}} & 0 & 0 & 0 & 0 \end{bmatrix},\\
I^{(4)} = \begin{bmatrix}0 & 0 & 0 & 0 & \textcolor{red}{\textbf{1}} & 0 & 0 & 0 & 0 \end{bmatrix},\\ \dots
\end{gather*}
We note that a drift may occur over a short number of time steps, and a minimum half-pixel shift necessarily occurs for odd numbers of time steps, and vibration here refers to the zero-mean ensemble or to an equal $L=R$ schedule over many layers.

In contrast, the least probable path is one in which the same operator is chosen at every layer, i.e., $L=N$ or $R=N$, in which case the signal drifts in one direction in a pure \textbf{translation mode}. The combined probability for either left or right motion is
\[
P(N,0) + P(0,N) = \alpha^N + (1 - \alpha)^N,
\]
which decays exponentially with \(N\) unless \(\alpha=0\) or \(\alpha=1\). To demonstrate, at each iteration convolve the impulse with a single difference orientation \(\nabla_{+}\) and apply ReLU:
\[
I^{(t)} = \max\bigl(0,\;I^{(t-1)} * \nabla_+\bigr)
\]
This causes the impulse to shift rightward:
\begin{gather*}
I^{(0)} = \begin{bmatrix}0 & 0 & \textcolor{red}{\textbf{1}} & 0 & 0\end{bmatrix},\\
I^{(1)} = \begin{bmatrix}0 & 0 & 0 & \textcolor{red}{\textbf{1}} & 0 & 0\end{bmatrix},\\
I^{(2)} = \begin{bmatrix}0 & 0 & 0 & 0 & \textcolor{red}{\textbf{1}} & 0 & 0\end{bmatrix},\\
I^{(3)} = \begin{bmatrix}0 & 0 & 0 & 0 & 0 & \textcolor{red}{\textbf{1}} & 0 & 0\end{bmatrix},\\
I^{(4)} = \begin{bmatrix}0 & 0 & 0 & 0 & 0 & 0 & \textcolor{red}{\textbf{1}} & 0 & 0\end{bmatrix},\\\dots
\end{gather*}

Pure vibration and translation thus represent extreme paths resulting from difference-operator application, in which information either vibrates in place or translates with a maximum net displacement in a single direction. In deep or randomly mixed-orientation networks, pure sustained vibrations or translations are rare, and small diffusive random walks around the origin become the typical behaviour. In practical scenarios such as conventional randomly initialized CNNs~\cite{lecun1989handwritten}, typical paths fall between pure vibration and translation and involve a random, small net displacement about the starting position, consistent with Brownian motion (random walk)~\cite{einstein1906theory}. The maximum displacement in the case of pure translation is limited by the discrete filter width to $\frac{Width-1}{2}$ per iteration (e.g.\ 0.5 pixels for gradient $\nabla=[-1,1]$), serving an analogous role to the maximum speed of light $c$ in special relativity.

We note that the diffusion, vibration and translation modes of information propagation resulting from rectified convolution are reminiscent of the physical transport of heat, momentum and mass. The following sections generalize these operators to higher image dimensionalities (i.e.\ 2D and 3D image space) via symmetry and to larger operator sizes via superposition and the discrete cosine spectrum, and demonstrate emergent phenomena.

%We note that in physics, the propagation of information (light) or particles (electrons, waves) also follows modes described above. For example, heat propagates by means of conduction (i.e. diffusion), convection (i.e. translation) and radiation as electrons translate and vibrate, etc. We will take some cases from physics, and we will try to draw similarities between them and the physics of image space, to continue to establish the mathematical framework in the image space that we have been building in terms of the 

% We note that in physics, the propagation of information (light) or particles (electrons, waves) also follows modes described above. For example, heat propagates by means of conduction (i.e. diffusion), convection (i.e. translation) and radiation as electrons translate and vibrate, etc. We will take some cases from physics, and we will try to draw similarities between them and the physics of image space, to continue to establish the mathematical framework in the image space that we have been building in terms of the fundamental $\Sigma$ and $\nabla$ operators.

\section{Elementary Operators}
\label{elementary}

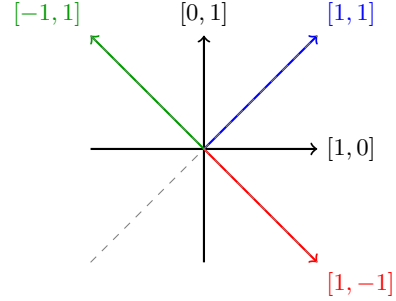
\begin{figure}[htb!]
  \centering
  \begin{tikzpicture}[scale=1]
    % Axes
    \draw[->, thick] (-1.5,0) -- (1.5,0) node[right] {$[1,0]$};
    \draw[->, thick] (0,-1.5) -- (0,1.5) node[above] {$[0,1]$};

    % Symmetric vector [1, 1]
    \draw[->, thick, blue] (0,0) -- (1.5,1.5) node[anchor=south west] {$[1,1]$};

    % Anti-symmetric vector [1, -1]
    \draw[->, thick, red] (0,0) -- (1.5,-1.5) node[anchor=north west] {$[1,-1]$};

    % Anti-symmetric vector [-1, 1]
    \draw[->, thick, green!60!black] (0,0) -- (-1.5,1.5) node[anchor=south east] {$[-1,1]$};

    % Dotted reflection lines
    \draw[dashed, gray] (-1.5,-1.5) -- (1.5,1.5);
    \draw[dashed, gray] (-1.5,1.5) -- (1.5,-1.5);

  \end{tikzpicture}
  \caption{Visualizing the geometry of symmetric sum $\Sigma = [1, 1]$ and antisymmetric gradient $\nabla_x = [1, -1]$ and $-\nabla_x = [-1, 1]$ operators, and axes defined by left $[1, 0]$ and right $[0, 1]$ positions.}
  \label{fig:filter_symmetry}
\end{figure}

The symmetric nature of the $\Sigma = [1,1]$ operator, the antisymmetric nature of the $\nabla = [-1, 1]$ operator, and their orthogonality in the space of 2-pixel filters may be visually appreciated from Figure~\ref{fig:filter_symmetry}. Note how rectified convolution constrains information to the positive orthant, centred about vector $[1,1]$ and bounded by vertical and horizontal axes representing information in right $[0,1]$ and left $[1,0]$ positions. In this section, we propose generalizing these elementary operators and their effects upon information to higher image-space dimensions and larger filter sizes using symmetry and superposition.

\subsection{Symmetry and Antisymmetry}
\label{symmetry}
Here we discuss symmetry and antisymmetry in mathematical functions. We use symmetric function and even function synonymously, and similarly an antisymmetric function and an odd function. We begin by describing left--right symmetry in 1D scalar functions, then radial symmetry and spherical symmetry and antisymmetry in 2D and 3D functions and discrete images.

\textbf{Symmetric Function:}  
A 1D function \( f: D \to R \) is said to be \emph{symmetric} if
\[
\forall x \in D, \quad f(x) = f(-x).
\]

\textbf{Antisymmetric Function:}  
A 1D function \( f: D \to R \) is said to be \emph{antisymmetric} if
\[
\forall x \in D, \quad f(-x) = -f(x).
\]

Now, any function can be uniquely decomposed into its symmetric and antisymmetric components. A more detailed discussion of the decomposition can be found in~\cite{smith2007mathematics}. The decomposition of \( f(x) \) into symmetric and antisymmetric components is given by:

\begin{equation}\label{eq:symm-anti-decomposition}
f(x) = f_s(x) + f_a(x),
\end{equation}

where symmetric \( f_s(x) \) and antisymmetric \( f_a(x) \) components are defined as
\begin{align}
    f_s(x) = \frac{f(x) + f(-x)}{2}, \notag \\
    f_a(x) = \frac{f(x) - f(-x)}{2}. \notag
\end{align}

These symmetric and antisymmetric functions are orthogonal, and therefore the squared norm of \( f \) satisfies the Pythagorean identity:
\begin{equation}\label{eq:pythagorean-symm-anti}
\|f\|^2 = \|f_s\|^2 + \|f_a\|^2
\end{equation}

\subsubsection*{Radial Symmetry for Functions}
The concept of symmetry may be extended to functions defined in multidimensional spaces, specifically radial symmetry here, as follows. Let \( f: \mathbb{R}^n \to \mathbb{R} \) be a function defined in \( n \)-dimensional space. We say that \( f \) is \textbf{radially symmetric} if its value depends only on the distance from a fixed point (usually the origin). Mathematically, this means
\[
f(x) = f(\|x\|),
\]
where \( \|x\| \) is the Euclidean norm (or distance) from the origin (or centre). In this case, the function value at any point is the same at all points equidistant from the centre.

In two dimensions, for example, \( f(x, y) \) is radially symmetric if
\[
f(x, y) = f\left(r\right) \implies f(x, y) = f\left(\sqrt{x^2 + y^2}\right),
\]
where \(r\) is the radius, meaning that the function's value is determined by the distance from the origin.

\subsubsection*{Radial Antisymmetry}
For a general function, a useful related construction is to subtract a radially symmetric (shell-averaged) component, yielding a remainder that is zero-mean on each radial shell. This remainder is orthogonal to the shell-averaged part, and generalizes the odd function defined by a coordinate inversion $x\mapsto -x$ in 1D space. For a function \(f(x, y) \) with radially symmetric component \(f_s(x, y)\), we write
\[
f_a(x, y) = f(x, y) - f_s(x, y),
\]
where by construction of shell averaging one has \(f_s \perp f_a\) in the Frobenius (pixelwise) inner product.
% \subsection{}
% For images, which are represented as matrices, we define an image \( I: \{0, \dots, n-1\} \times \{0, \dots, n-1\} \to \mathbb{R} \) (for a grayscale image) to be \textbf{radially symmetric} with respect to its center if the pixel values depend only on their distance from the center of the image. Formally, for an image with a center located at \( c_1 = c_2 = \frac{n-1}{2} \), we have:

% \begin{equation}\label{eq:constructing-symm-component}
% I(x, y) = f\left(\sqrt{(x - c_1)^2 + (y - c_2)^2}\right)
% \end{equation}
% where \( (x, y) \) are the pixel coordinates. This means that all points equidistant from the center of the image have the same value.

\subsection{Radial Symmetry in Images}
\label{app:radial3d}

For images, which are represented as matrices, we define a 3D scalar image \(I\colon\{0,\dots,N-1\}^3\to\mathbb{R}\) as a cubic volume.  Its centre is
\[
(p,p,p),\quad p=\frac{N-1}{2}.
\]
Defining the radial distance
\[
r_{x y z}
=\sqrt{(x-p)^2+(y-p)^2+(z-p)^2},
\]
we group all voxels with the same \(r\) into a \emph{shell}, and define the \emph{radially symmetric} component by averaging:
\[
I_s(x,y,z)
= \frac{1}{|\mathcal{S}_r|}\sum_{(x',y',z')\in \mathcal{S}_r}
  I(x',y',z'),
\]
where $\mathcal{S}_r$ is the set of all points at radial distance $r$,
\[\mathcal{S}_r=\{(x',y',z')\,:\,r_{x'y'z'}=r\}.\]

In 2D, \(|\mathcal{S}_r|\) \(\leq\) 8, and in 3D \(|\mathcal{S}_r|\) \(\leq\) 24.

Thus, radially symmetric images exhibit circular or spherical symmetry, where the pixel intensities vary as a function of the distance from the centre but not with direction (angle).

The \emph{antisymmetric} (shell-demeaned) remainder is
\[
I_a = I - I_s.
\]

In 3D the natural symmetry group of a cube is the octahedral (cubic) group \(O_h\), a finite subgroup of \(\mathrm{SO}(3)\) that preserves cubic symmetry~\cite{altmann1986rotations,dresselhaus2008group}. We can alternatively invoke the full rotation group \(\mathrm{SO}(3)\) for continuous radial invariance~\cite{hall2003lie}. A worked example of a $3 \times 3$ matrix can be found in Appendix~\ref{app:decomp-example}.

% \subsection{Decomposing a Function into Symmetric and Anti-symmetric Components}
% Finally, a function \( f(x) \) (or an image \( I(x, y) \)) can be decomposed into symmetric and anti-symmetric components, where the symmetric part depends only on the distance from the center, and the anti-symmetric part changes sign under reflection through the center, and can be obtained by simply subtracting the symmetric part from the total function. Referring to \eqref{eq:symm-anti-decomposition}, this decomposition can be expressed as:

% \[
% f(x) = f_s(x) + f_a(x)
% \]
% where \( f_s(x) \) is the radially symmetric component and \( f_a(x) \) is the radially anti-symmetric component.

% The symmetric part captures the circular symmetry, while the anti-symmetric part captures the asymmetry across the origin or center.

%\section{Operators in Image Space}
\subsection{Operators in Image Space}

The notions of symmetry and antisymmetry, sum and gradients must be generalized to larger filter sizes and image dimensions for practical contexts. For example, CNN-based image classification~\cite{lecun2015deep} typically adopts small real-valued filters of odd side length, i.e.\ size $3 \times 3$ pixels.
The minimal sum $\Sigma=[1,1]$ and gradient $\nabla=[-1,1]$ operators are related in form to a variety of contexts and transforms, e.g.\ the Hadamard quantum logic gate~\cite{williams2010explorations}, hierarchical wavelets~\cite{haar1909theorie,gabor1946theory}, Gaussian derivatives~\cite{lindeberg2013scale} and the discrete cosine transform (DCT)~\cite{britanak2010discrete} we adopt here. They resemble oriented gradient-like functions emerging as sparse codes or principal components of natural images and CNN filters~\cite{olshausen1996emergence,fukuzaki2022principal} or found in simple-cell receptive fields of the mammalian visual cortex~\cite{hubel1959receptive}, with gradients sampled over equal angular increments in traditional image descriptors~\cite{lowe2004distinctive} or equivariant neural network filtering~\cite{satorras2021n}.

Our work here analyzes the effect of such operators under rectified convolution, introduced relatively recently~\cite{nair2010rectified} and popularized in the context of neural networks~\cite{krizhevsky2012imagenet}. We adopt the DCT as a fixed, computationally efficient set of basis functions, used ubiquitously for example in JPEG image compression. Although DCT kernels are not explicitly localized in space as wavelets, when applied blockwise they produce localized, oriented responses that also capture edges and blobs at multiple scales. This paper discusses three such basis components of the DCT decomposition, most notably the $\Sigma$ (sum-like) and $\nabla_x, \nabla_y$ (gradient-like) components. Prior work has shown these to represent greater than $92\%$ of the classification accuracy in the CNN settings studied in~\cite{frija2025mechanics}, motivating their practical value.

\begin{figure}[h]
    \centering
    \includegraphics[width=0.25\textwidth]{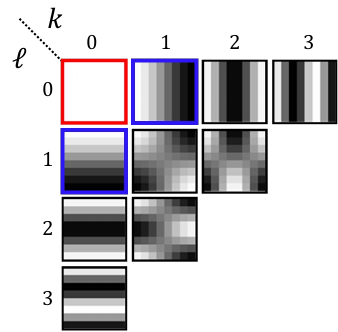}
    \caption{The 10 DCT components corresponding to mixed partial derivative operators of up to third order in 2D image space. The red box shows the DC or sum $\Sigma$ component (all white), the two blue boxes show the horizontal and vertical gradient components $\nabla_x,\nabla_y$ (black to white).}
    \label{fig:dct}
\end{figure}

The DCT is a set of real-valued basis functions that can be represented as an $N \times N$-pixel matrix in 2D image space as illustrated in Figure~\ref{fig:dct}, or as an \(N\times N \times N\)-voxel cube in 3D volumetric image space, with coefficients $C_{k\ell m}$ defined as follows:

\begin{multline}
C_{k\ell m}
= \alpha_k\,\alpha_\ell\,\alpha_m
  \sum_{i=0}^{N-1}\sum_{j=0}^{N-1}\sum_{p=0}^{N-1}
    I_{ijp}\,
    \cos\!\Bigl(\tfrac{\pi(2i+1)k}{2N}\Bigr) \\
  \times
    \cos\!\Bigl(\tfrac{\pi(2j+1)\ell}{2N}\Bigr)
    \cos\!\Bigl(\tfrac{\pi(2p+1)m}{2N}\Bigr)
\end{multline}

where
\[
\alpha_0 = \sqrt{\tfrac{1}{N}},\quad
\alpha_{>0} = \sqrt{\tfrac{2}{N}}
\]
Just as JPEG compression discards high‐frequency coefficients to compress 2D images, a 3D codec will quantize or zero‐out coefficients \(C_{k\ell m}\) with large \((k,\ell,m)\) to compress a volume. By viewing CNN filters in terms of these 3D DCT basis functions, we gain direct insight into frequency localization and potentially leverage JPEG‐style quantization strategies in learned networks.

% \begin{figure}[h]
%     \centering
%     \includegraphics[width=0.3\textwidth]{figures/DCT_4x4_triangle.png}
%     \caption{DCT components, showing the $16=4\times 4$ low frequency basis functions for $8\times8$ pixel-sized images. Red boxes highlight the components associated with the DC or sum $\Sigma$ (all white) and gradient $\nabla_x,\nabla_y$ (half black, half white) operators.}
%     \label{fig:dct}
% \end{figure}

In image space, an operator applied to an image can be interpreted as analogous to a matched filter in signal processing, detecting specific patterns or ``signals'' against a background of noise~\cite{turin1960introduction}. More specifically, each CNN filter (or DCT basis) acts like a template that ``matches'' particular spatial features, as Turin formalized for optimal detection in the presence of Gaussian noise. The three components referenced above are the three highlighted components in Figure~\ref{fig:dct}. The first component $(\ell,k)=(0,0)$ in the top left corner (all white) can be treated as the $\Sigma$ component, with a similar constant integration effect. The other two components $(\ell,k)=(1,0), (0,1)$ (half black, half white) have the effect of horizontal $\nabla_x$ or vertical gradient $\nabla_y$ filters, that may be mixed linearly to represent gradients at intermediate angles~\cite{freeman1991design}. The DCT is known for its high ``energy compaction'' property of concentrating the majority of signal energy into low-frequency components. For example, natural image structure such as contours is more efficiently encoded with antiperiodic DCT components (e.g.\ gradients) than periodic Fourier components. We hypothesize that these fundamental components can explain the primary modes of information flow in 2D and 3D image space, analogously to the $\Sigma=[1, 1]$ and $\nabla=[-1, 1]$ operators in 1D space.

%which improves compression and expressing einsteins field equations~\cite{christensen2007haar}.

\section{Emergent Phenomena}

In this section, we discuss emergent phenomena in image space relating to sum $\Sigma$ and gradient $\nabla$ operators, including several experiments that help to understand these. The first involves the impulse response of operators, which produces displacement behaviour analogous to a Lorentz transform. The second involves the accumulation of information over scale and time, where spherical bubble-like and droplet-like structures emerge as topological critical points in an image diffusion process, sharing Laplacian structure with the heat equation and, by mathematical analogy, with the free Schr\"odinger equation and with Friedmann expansion parameterized by a time-dependent cosmic scale factor $a(t)$.

\subsection{Lorentz-like Displacement in Image Space}

This section demonstrates how discrete filtering with rectified convolution as in neural networks can lead to displacement behaviour reminiscent of special relativity. We begin with the relativistic energy--momentum relation, which is central to modern physics and is defined as:
\begin{equation}\label{eq:energy-momentum}
E^2 = (mc^2)^2 + (pc)^2 = (\gamma mc^2)^2,
\end{equation}
where in the first equality, \(E\) is the total energy of a particle with rest mass \(m\) and momentum \(p\), and $c$ is the light speed constant. In the second equality, the energy of a particle moving at speed $v$ is expressed in terms of the Lorentz gamma factor $\gamma$, as described in detail in Appendix~\ref{app:lorentz} and defined as:
\begin{equation}\label{eq:beta2}
\beta = \frac{v}{c}, \quad
    \gamma = \frac{1}{\sqrt{1 - \beta^2}},    
\end{equation}
where \(\beta\) represents the ratio of particle speed $v$ relative to the maximum \(c\). Rearranging Equations~\eqref{eq:energy-momentum} and~\eqref{eq:beta2} reveals that \(\beta\) also determines the ratio of momentum $pc$ to rest energy $mc^2$:
\begin{align}\label{eq:beta1}
    \frac{\beta}{\sqrt{1-\beta^2}} &= \frac{pc}{mc^2}.
\end{align}

In quantum mechanics, rest mass $m$ is typically treated as a real fixed quantity rather than an operator, while the linear momentum operator is defined as proportional to the gradient $\hat{p}=-i\hbar\nabla$ with reduced Planck constant $\hbar$. In CNNs, filtering with a purely symmetric component $f=f_s$ (e.g.\ sum $f_s=\Sigma$) has an isotropic point-spread effect upon information (e.g.\ diffusion) that preserves its centre of mass (CoM), while filtering with an antisymmetric component $f=f_a$ (e.g.\ gradient $f_a=\nabla$) generally induces a directional shift in the CoM. In this way, symmetric and antisymmetric filter components bear a superficial similarity to rest energy $mc^2$ and momentum term $pc$ in the linear Dirac equation from Appendix~\ref{app:dirac} and the quadratic energy--momentum relation in Equation~\eqref{eq:energy-momentum}.

We thus hypothesize the following analogy between the energy components of a filter in image space and a relativistic particle in physical space:
\begin{equation}\label{eq:equivFE}
    \|f\| \equiv E \qquad
    \|f_a\| \equiv pc \qquad
    \|f_s\| \equiv mc^2
\end{equation}

To test this hypothesis, we perform convolution experiments involving a test pattern and a filter $f$ defined by symmetric and antisymmetric components mixed according to ratio $\beta$ in Equation~\eqref{eq:beta2} as follows:
\begin{align}
    f = \|f\|\beta \hat{f_a} + \|f\|\sqrt{1-\beta^2}\hat{f_s},
    \label{eq:beta_mixing}
\end{align}
where in Equation~\eqref{eq:beta_mixing}, $\hat{f_s}=\frac{f_s}{\|f_s\|}$ and $\hat{f_a}=\frac{f_a}{\|f_a\|}$ are symmetric and antisymmetric components normalized to unit length. The displacement of a test pattern centre of mass may then be measured for filters at various values of mixing coefficient $\beta = \frac{\|f_a\|}{\|f\|}$, and compared to the expected displacement given a Lorentz transform with parameter $\beta = \frac{v}{c}$.

\subsubsection{Lorentz-like Motion from Rectified Convolution}

Here we demonstrate that rectified convolution yields Lorentz-like CoM motion for filters mixed according to $\beta$ as in Equation~\eqref{eq:beta_mixing}. A single-pixel impulse test pattern is convolved 10 times with a filter mixing sum $\Sigma$ and horizontal gradient $\nabla_x$ components, for five values of mixing parameter $\beta^2 = \{0,.25,.5,.75,1\}$. At each time step, the centre of mass $\mu_x$ and standard deviation $\sigma_x$ of the resulting image pattern are computed along the central pixel row $y=0$, where $\mu_x$ represents the mean displacement in the horizontal direction $x$ and may be used to estimate velocity in the image plane. 

\begin{figure*}[ht!]
    \includegraphics[width=.8\linewidth]{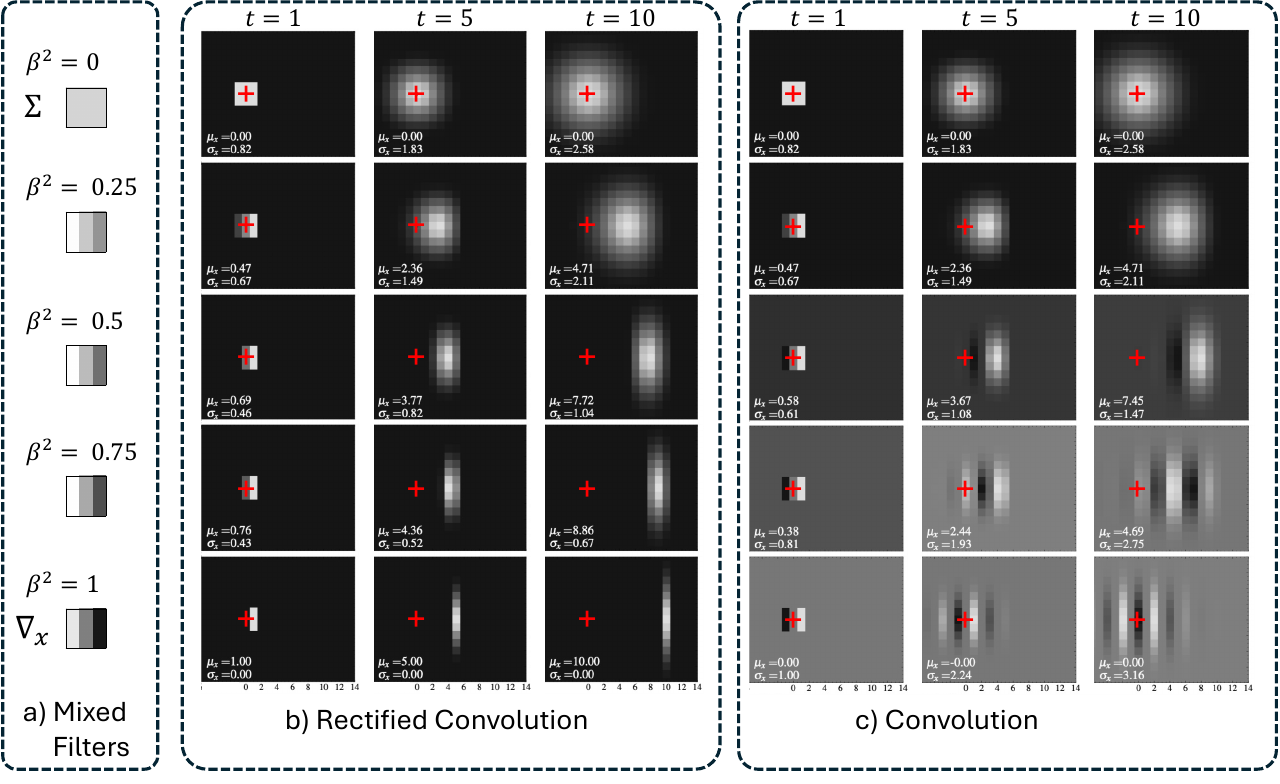}
  \caption{Comparing repeated b) rectified convolution vs c) standard convolution of an impulse pattern for a) five different mixed filters at iterations $t=1,5,10$ (columns).}
  \label{fig:prop_diag_rect_vs_norect}
\end{figure*}

\begin{figure}[htb!]
    \includegraphics[width=.8\linewidth]{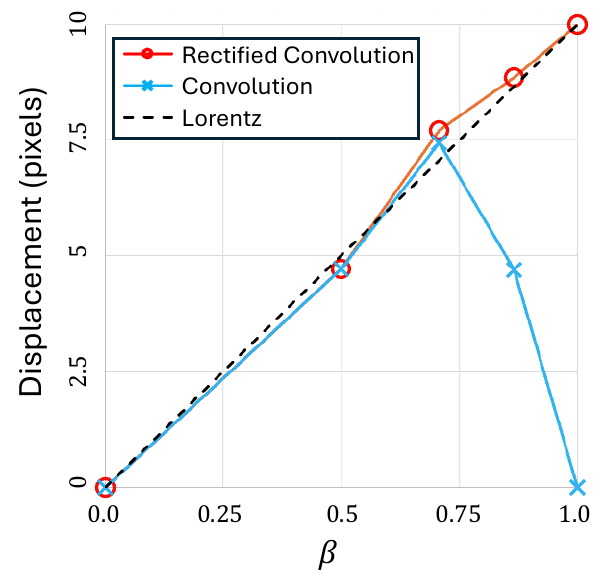}
  \caption{The centre of mass displacement (vertical axis) vs mixing coefficient $\beta$ (horizontal axis) for ten iterations of rectified convolution (red), standard convolution (blue) and the Lorentz transform (dashed line). The displacement for rectified convolution follows the Lorentz transform closely, but for standard convolution diverges for values of $\beta > 0.75$.}
  \label{fig:prop_graph}
\end{figure}

The result of convolution with rectification is shown in Figure~\ref{fig:prop_diag_rect_vs_norect} b), where the centre of mass is stationary with zero displacement $\mu_x = 0$ for pure diffusion at $\beta^2=0$ and maximal at $\mu_x = 10$ pixels for pure translation $\beta^2=1$. The result for standard linear convolution without rectification shown in Figure~\ref{fig:prop_diag_rect_vs_norect} c) is identical for values of $\beta^2 < 0.5$ where the result of filtering is positively valued; however, for $\beta^2 \ge 0.5$ a wave packet with positive and negative values emerges and group displacement falls to zero for $\beta^2=1$. The graph in Figure~\ref{fig:prop_graph} shows that the displacement for convolution with rectification (red) follows a Lorentz-transform curve (dashed line) closely across the range of $\beta$, whereas it drops to zero for convolution without rectification. To our knowledge, this Lorentz-like displacement under rectified convolution has not been previously reported.

\subsubsection{Diffusion, Vibration and Translation}

\begin{figure*}[ht!]
    \includegraphics[width=.7\linewidth]{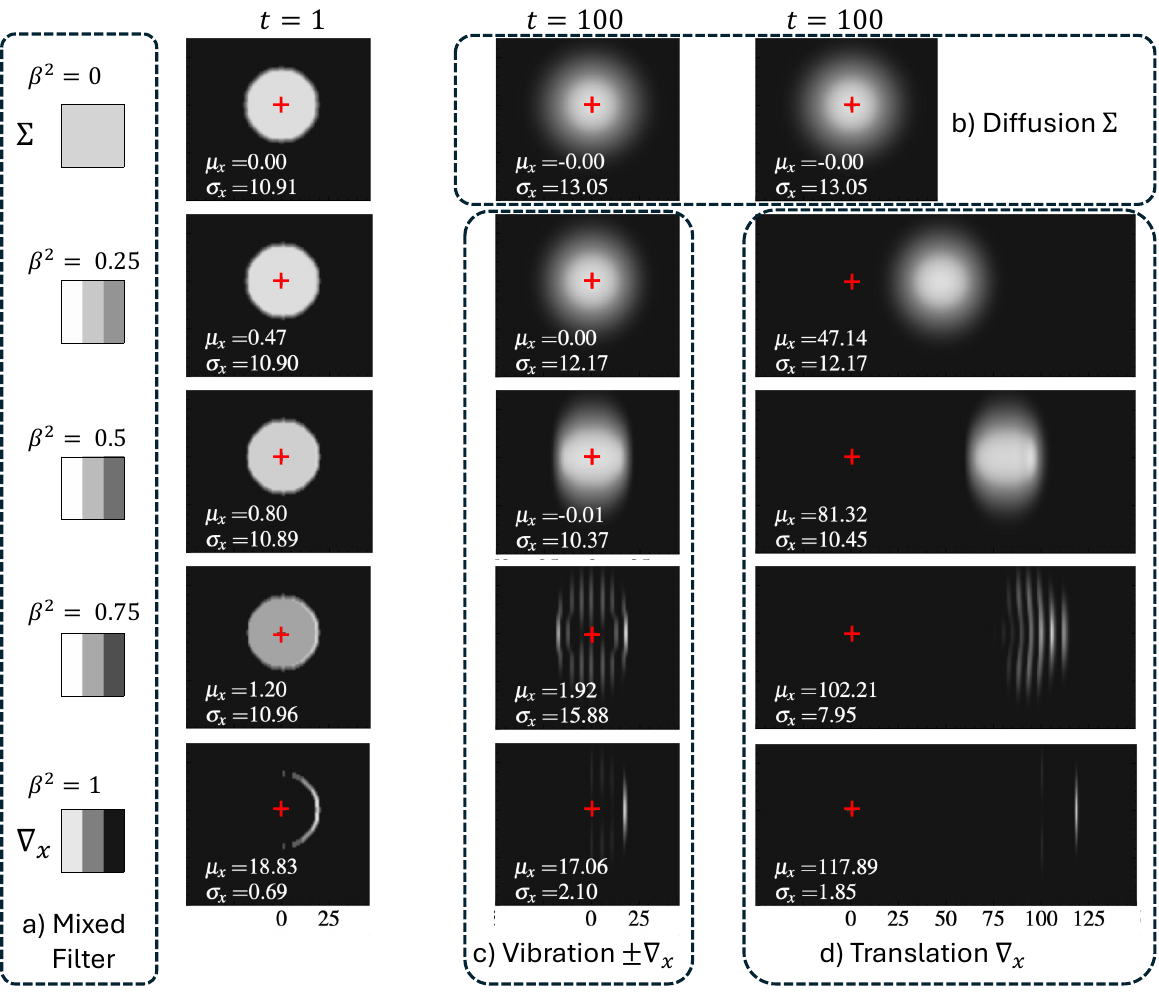}
  \caption{Demonstrating b) diffusion c) vibration and d) translation modes of propagation resulting from repeated {\bf rectified convolution} of a circle test pattern with five different mixed filters a) at iterations $t=1,100$ (columns).}
  \label{fig:prop_diag_relu}
\end{figure*}

Here we demonstrate further Lorentz-transform-like behaviour resulting from rectified convolution, including primary diffusion, vibration and translation modes of propagation as shown in Figure~\ref{fig:prop_diag_relu}. A circular test image pattern (of radius 19 pixels) is convolved 100 times with a $3 \times 3$-pixel filter mixing sum $\Sigma$ and horizontal gradient $\nabla_x$ components according to five values of mixing ratio $\beta^2$ as shown in Figure~\ref{fig:prop_diag_relu} a). The result of rectified convolution after the first time step $t=1$ and after $t=100$ steps is shown in the columns of Figure~\ref{fig:prop_diag_relu}.

Diffusion arises in the case of $\beta^2 = 0$ when the filter is pure sum $\Sigma$, and is associated with a stationary centre of mass $\mu_x = 0$ and isotropically increasing standard deviation $\sigma_x$ due to filter symmetry, as shown in Figure~\ref{fig:prop_diag_relu} b). Motion results when the filter contains a non-zero gradient component $\nabla_x$ (i.e.\ $\beta^2 > 0$), and emerges as vibration or translation as shown in Figure~\ref{fig:prop_diag_relu} c) and d).  

Vibration arises when the horizontal gradient component is applied in alternating left and right directions $\pm \nabla_x$ with each iteration $t$ as shown in Figure~\ref{fig:prop_diag_relu} c). Note that the case of $\beta^2 = 0$ (top row of Figure~\ref{fig:prop_diag_relu}) corresponds to pure diffusion and the test pattern centre of mass $\mu_x=0$ remains stationary. In the case of $\beta^2 \le 0.5$, the pattern centre of mass vibrates about $\mu_x=0$ and its standard deviation $\sigma_x$ contracts in the horizontal direction of motion. In the case of $\beta^2 = 1$, the bulk of the pattern disappears, and a wave front of approximately a single pixel width vibrates about the location of the right contour of the test pattern.

Translation arises when the horizontal gradient component $\nabla_x$ is applied in a single direction as shown in Figure~\ref{fig:prop_diag_relu} d). Note again that the case of $\beta^2 = 0$ is pure diffusion and independent of the gradient. For $\beta^2 = 0.5$, the pattern both diffuses and translates horizontally, its standard deviation $\sigma_x$ decreasing in the horizontal direction of travel, a foreshortening effect reminiscent of length contraction associated with a rapidly moving object under a Lorentz transform. For $\beta^2 = 1$, the pattern bulk disappears, and a single-pixel ``wave front'' travels from the right contour of the circle to a maximum distance of 100 pixels.

Together, these experiments demonstrate the effect of repeated filtering with filters mixing sum $\Sigma$ and gradient $\nabla$ components according to the $\beta$ ratio, which is analogous to what might be expected of Lorentz transforms mixing mass and momentum components according to $\beta = \frac{v}{c}=\frac{pc}{E}$. A centre-of-mass displacement occurs with each iteration due to the gradient $\nabla$ component, where the maximum displacement is determined by the filter width, e.g.\ a single pixel $\frac{Width-1}{2}=1$ per iteration for $3 \times 3$ filters, and occurs when $\beta^2=1$. A net displacement generally emerges based on the number of right vs.\ left gradient operations. At one extreme, equally alternating between right and left gradient directions $\pm \nabla_x$ causes information to vibrate in place, similarly to a bound oscillatory degree of freedom as in Figure~\ref{fig:prop_diag_relu} c). At another extreme, a single gradient direction $\nabla_x$ causes information to translate in a single direction up to a maximum distance, similarly to a signal propagating at a finite maximum speed as in Figure~\ref{fig:prop_diag_relu} d).

\subsection{Diffusion and Scale-Invariant Image Structure}

In this section we discuss scale-invariant features that emerge following multiple layers of rectified convolution. The central limit theorem ensures that repeated application of IID binomial filtering converges to Gaussian filtering dominated by the $\Sigma$ component, while the effect of components such as gradients $\nabla$ averages out at large scales. The Gaussian filtering process is analogous to heat conduction or particle diffusion~\cite{einstein1906theory}, and scale-invariant features here are emergent spherical structures exhibiting the topology of either bubbles or droplets. They may be detected in the image in a manner invariant to similarity transforms of the image (i.e. scaling, rotation and translation), and thus characterized by canonical coordinates independently of image transforms. These so-called SIFT features have been widely used in 2D computer vision~\cite{lowe2004distinctive} and 3D medical image analysis~\cite{toews2013efficient}, but are not widely known in the physical sciences. Here we discuss diffusion in physical and image space, and critical points of the Laplacian-of-Gaussian scale-space, which are observed in the experiments of the following section.

\subsubsection*{Governing PDE in Physics and Images (3D)}

In physics, the 3D heat (diffusion) equation on a rectangular domain is
\begin{equation}\label{eq:heat-pde-3d}
\frac{\partial u}{\partial t}
= \kappa^2\bigl(u_{xx}+u_{yy}+u_{zz}\bigr)\,
\end{equation}
where $u$ is the temperature density with Dirichlet boundary conditions and initial condition \(u(x,y,z,0)=f(x,y,z)\), and $\kappa^2$ is the thermal diffusivity constant. The full derivation and its solution via separation of variables are presented in Appendix~\ref{app:Diffusion}. The analogy between this physical PDE and the image–scale PDE is noted in Jan J.\ Koenderink’s classic “The Structure of Images” \cite{koenderink1984structure} and in the topological analysis of scale‐space by Florack and Kuijper \cite{florack2000topological}.

In image space, the diffusion process is parametrized not by physical time \(t\), but by the scale \(\sigma\), which is related to time via \(t = \sigma^2\). The image \(I\colon\mathbb{R}^3\to\mathbb{R}\) evolves in scale as:
\begin{equation}\label{eq:scale-pde-3d}
\frac{\partial I}{\partial \sigma}
= 2D\,\sigma\,\bigl(I_{xx}+I_{yy}+I_{zz}\bigr)
\end{equation}
conditioned by the initial image $I_0$:
\[
I(x,y,z,0) = I_0(x,y,z)
\]
where \(D > 0\) is the isotropic diffusion constant. This version accounts for the transformation \(t = \sigma^2\), and makes explicit the correspondence between physical diffusion and scale-space evolution in image analysis.

\subsubsection*{Solution via 3D Gaussian Smoothing}

The fundamental solution of the 3D diffusion equation is a convolution with the isotropic 3D Gaussian kernel. Using the time–scale relation \(t = \sigma^2\), the solution becomes:
\begin{equation}\label{eq:gaussian-scale-space-3d}
I(x,y,z,\sigma)
= G_3(x,y,z;\sigma)\;\ast\;I_0(x,y,z)
\end{equation}
\[
G_3(x,y,z;\sigma)
= \frac{1}{(4\pi D \sigma^2)^{3/2}}
  \exp\!\Bigl(-\frac{x^2+y^2+z^2}{4D\sigma^2}\Bigr)
\]
where \(\sigma\) is the standard deviation of the Gaussian kernel and determines the scale level in the image.

\subsubsection*{3D Laplacian and Difference-of-Gaussian Approximation}

The full scale-normalized Laplacian-of-Gaussian (LoG) in 3D is as follows:
\[
\LoG = \mathcal{L}\{I\}(\sigma)\;=\;\sigma^2\,\nabla^2 I(\,\cdot\,,\sigma).
\]
Rather than the LoG, SIFT uses the Difference-of-Gaussian (DoG) approximation:
\begin{align}
\DoG(x,y,z;\sigma)
&= I(x,y,z;k\sigma)\;-\;I(x,y,z;\sigma)
\label{eq:dog-def-3d}\\
&\approx (k-1)\,\sigma^2\,\nabla^2 I(x,y,z;\sigma)
\notag
\end{align}
valid for small \(k-1\).  Hence  
\(\DoG\propto\sigma^2\nabla^2I\),  
and the extrema of \(\DoG(x,y,z,\sigma)\) mark 3D scale-invariant keypoints discussed later. Parameter $k>1$ defines the geometric sampling rate of the scale parameter $\sigma_t=k\sigma_{t-1}$, ensuring scale-invariance of features detected, and is defined by the number of samples $N$ per octave or doubling $k=2^{\frac{1}{N}}$. In experiments we use $k=2^{\frac{1}{3}}$ recommended as a compromise between precision of detection vs data efficiency for natural photographs~\cite{lowe2004distinctive}. This parameter is discussed further in Section~D.

\subsubsection*{3D Hessian Matrix and the Laplacian}

The Hessian of a twice–differentiable volume \(I(x,y,z)\) is the \(3\times3\) matrix
\[
H I
= \begin{pmatrix}
I_{xx} & I_{xy} & I_{xz}\\
I_{yx} & I_{yy} & I_{yz}\\
I_{zx} & I_{zy} & I_{zz}
\end{pmatrix}
\]
Its trace recovers the Laplacian,
\begin{equation}\label{eq:laplacian-trace-3d}
\operatorname{tr}(H I)
= I_{xx}+I_{yy}+I_{zz}
= \nabla^2 I
\end{equation}
which drives both the physical diffusion \eqref{eq:heat-pde-3d} and the image scale–space PDE \eqref{eq:scale-pde-3d}. Meanwhile, determinants and principal minors of \(H\) underlie 3D blob detectors~\cite{harris1988combined,surf}, providing complementary image keypoint structure.

Concretely, a SIFT keypoint occurs wherever all partial derivatives of the LoG over space and scale are zero:
\[
\frac{\partial}{\partial x_i}\bigl(\LoG\ \bigr)
=0,\quad
\frac{\partial}{\partial \sigma}\bigl(\LoG\ \bigr)
=0.
\]
These may be described as Morse critical points~\cite{damon1995local}, i.e.\ points where the gradient of the LoG vanishes, but where the Hessian is non-degenerate~\cite{morse1929foundations}. Maxima or minima of the scale-normalized $\LoG$ response correspond to locally convex or concave structure for which the eigenvalues of the $3\times3$ Hessian are either all positive $(+,+,+)$ or all negative $(-,-,-)$, i.e.\ points of Morse index 0 or index 3. We refer to these as ``bubbles'' or ``droplets'', as they correspond to local minima or maxima of the density, analogously to vapour bubbles in water or water droplets in air, respectively. They may be assigned a topological charge based on the eigenvalue product (i.e.\ the determinant), which is positive for index~0 $(+,+,+)$ and negative for index~3 $(-,-,-)$.

The construction of the scale-space from fine-to-coarse scale via Gaussian blurring with exponentially increasing variance $\sigma^2 \propto k_BT$ is analogous to ``heating'' the image. A large number of fine-scale bubbles form at nucleation sites throughout the image, and progressively merge into a hierarchy of fewer and larger bubbles, until a single one exists. Neighbouring bubbles are naturally separated by saddle points, i.e.\ Morse critical points of index 1 or index 2 with Hessian matrices having one or two negative eigenvalues $(+,+,-)$ or $(+,-,-)$, analogously to Lagrange points in a gravitational field. Neighbouring bubbles merge as the scale $\sigma$ of blurring increases, annihilating with saddle points as described by catastrophe theory~\cite{thom1974stabilite}. The majority of features organize into a fine-to-coarse hierarchy over scale $\sigma$ as demonstrated in 2D images~\cite{florack2000topological}; however, new critical points may occasionally form~\cite{damon1995local}, producing loop structures as demonstrated in~\cite{kuijper2004relevance}. 

%Our work here focuses on the bubbles themselves, rather than their topological hierarchy.

%The Gaussian scale-space is known to be a unique filter satisfying 5 scale-space axioms~\cite{lindeberg2013scale}, including the non-creation or enhancement. 

\subsubsection*{Bridging Physics, Diffusion, and SIFT in 3D}

Thus, the classical heat-equation solution in 3D, its Gaussian kernel, the Laplacian operator, and modern 3D keypoint detectors (DoG and Hessian-determinant) all emerge from one coherent diffusion framework. To summarise:

\begin{enumerate}
  \item The solution \eqref{eq:gaussian-scale-space-3d} of the 3D scale–space PDE \eqref{eq:scale-pde-3d} is \emph{exactly} the Gaussian smoothing used by SIFT in volumetric data.
  \item The 3D DoG \eqref{eq:dog-def-3d} provides a \emph{finite‐difference} approximation to the 3D Laplacian, and its extrema across \((x,y,z,\sigma)\) are the SIFT keypoints (the “bubbles” and “droplets”).
  \item The 3×3 Hessian \(H I\) unifies both approaches: its trace \eqref{eq:laplacian-trace-3d} recovers the Laplacian that drives diffusion, while its determinant (and principal minors) underlies Hessian‐based blob detectors.
\end{enumerate}

%The Gaussian scale-space is known to be a unique filter satisfying 5 scale-space axioms~\cite{lindeberg2013scale}, including the non-creation or enhancement. 

\subsection{Scale-Invariant Diffusion Bubble Observations}

Here we present experiments evolving different types of images in the Gaussian scale-space and observing the emergent scale-invariant critical-point structure. The novelty here is in showing how phenomena across a wide range of physical scales may be represented as a hierarchy of bubble-like and droplet-like structures in information space, here displayed as either blue or yellow coloured spheres overlaying the images. We investigate electron density (DFT) images of sugar molecules and silicon crystals, 3D magnetic resonance brain scans of primates, and cosmological images including a galaxy simulation and the cosmic microwave background (CMB). We note that other algorithms also extract stable, invariant features, such as invariant scattering convolution networks~\cite{bruna2013invariant}, which leverage wavelet transforms without extensive training. We employed the 3D SIFT algorithm as described in~\cite{toews2013efficient}. The simplicity of SIFT allowed us to connect these observations to information mechanics and diffusion bubbles. SIFT achieves rotation invariance by assigning a dominant orientation to each keypoint based on local gradient distributions, i.e. a local coordinate reference frame, analogous to how humans recognize key features of real-world objects under rotation, as described by Shepard et al.~\cite{shepard1971mental}.
 
\begin{figure}[h]
    \centering
    \includegraphics[width=0.46\textwidth]{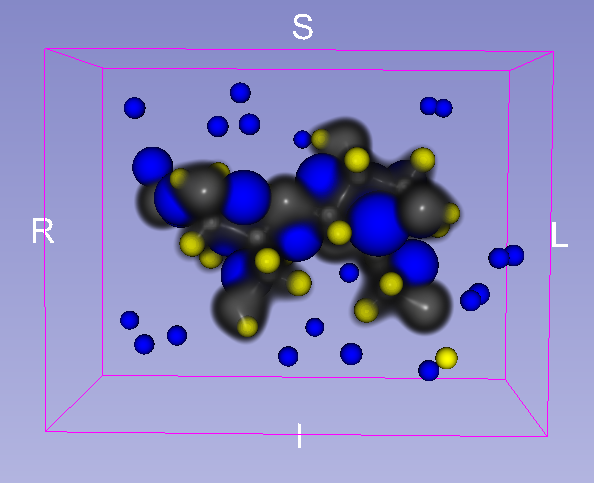}
    \caption{3D SIFT on a sucrose molecule \(C_{12}H_{22}O_{11}\).}
    \label{fig:sucrose-sift}
\end{figure}

In the case of the sucrose molecule in Figure~\ref{fig:sucrose-sift}, the yellow spheres represent droplets and correspond to atom locations; here there are 23 non-hydrogen atoms including 12 carbon and 11 oxygen atoms. The blue bubbles in the same case correspond to voids in the molecular space, reminiscent of atomic bond structure. We note that the Laplacian of electron density images has been studied at the molecular scale~\cite{bader1985atoms,popelier2000full,anderson2010ambiguous}, however not including a variable scale parameter as in the SIFT method leading to a hierarchy of droplet-like and bubble-like structure. These features may be useful in molecular indexing and matching, for example as molecular motifs to identify potential binding sites. Popelier also notes that schemes obtained by analysis of the Laplacian can be used for convenient classification purposes and in chemical databases.

\begin{figure}[h]
    \centering
    \includegraphics[width=0.46\textwidth]{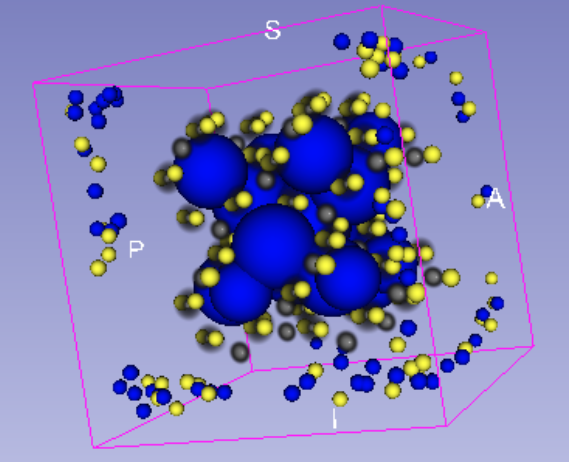}
    \caption{3D SIFT on a boron-doped silicon crystal structure, with 72 atoms.}
    \label{fig:silicon-sift}
\end{figure}

\begin{figure}[h]
    \centering
    \begin{tabular}{cc}
       \includegraphics[width=0.17\textwidth]{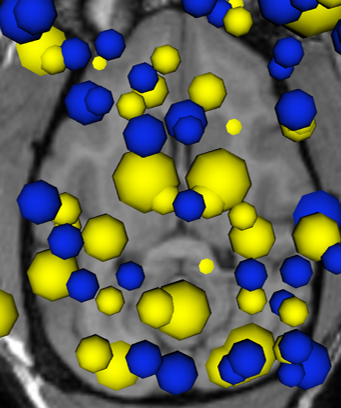}  &  \includegraphics[width=0.28\textwidth]{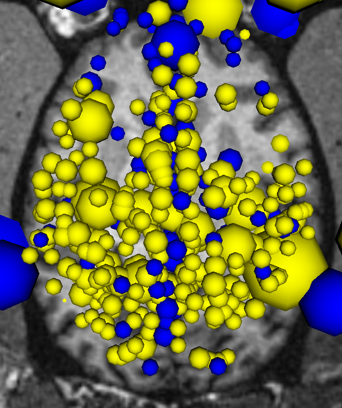} \\
       a) Macaque & b) Chimpanzee \\
\end{tabular}
\\
\includegraphics[width=0.4\textwidth]{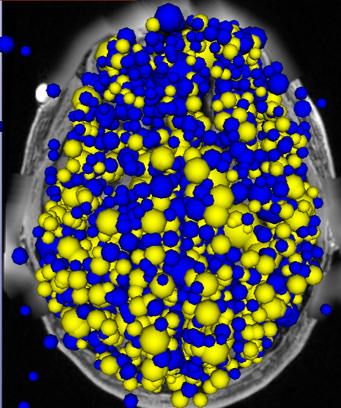} \\
c) Human

    \caption{3D SIFT on brain MRI of three primate species: a) macaque, b) chimpanzee and c) human.}
    \label{fig:macaque-sift}
\end{figure}

For the doped silicon crystal in Figure~\ref{fig:silicon-sift} we can see bubble formation similar to that of the sucrose molecule, i.e.\ atoms are identified as positive features (yellow), and inter-atomic spaces are associated with hole-like negative features (blue). For the brain MRI scans of primates as shown in Figure~\ref{fig:macaque-sift}, we have observed various symmetrical patterns of these features emerging from the diffusion process (3D SIFT); in these examples, positive features (yellow) tend to coincide with white-matter and sulcal regions, and negative features (blue) with ventricles and extra-cerebral spaces. Certain regions of the brain have a specific feature signature which can again help in more efficient indexing and matching, building on recent advances in pairwise neuroimage analysis that combine 3D keypoint sets with Jaccard-based similarity metrics~\cite{chauvin2021efficient} similarity metrics for image registration or alignment.

\begin{figure}[h]
    \centering
    \includegraphics[width=0.5\textwidth]{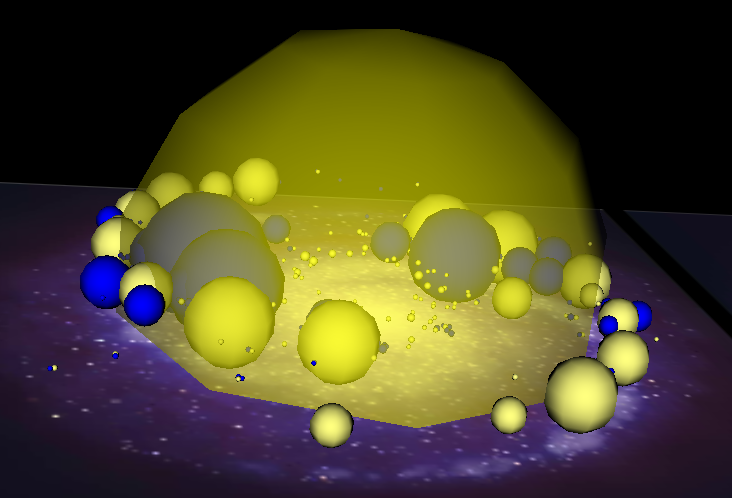}
    \caption{SIFT bubbles from a 2D galactic simulation profile.}
    \label{fig:galaxy-sift}
\end{figure}

\begin{figure}[h]
    \centering
    \includegraphics[width=0.5\textwidth]{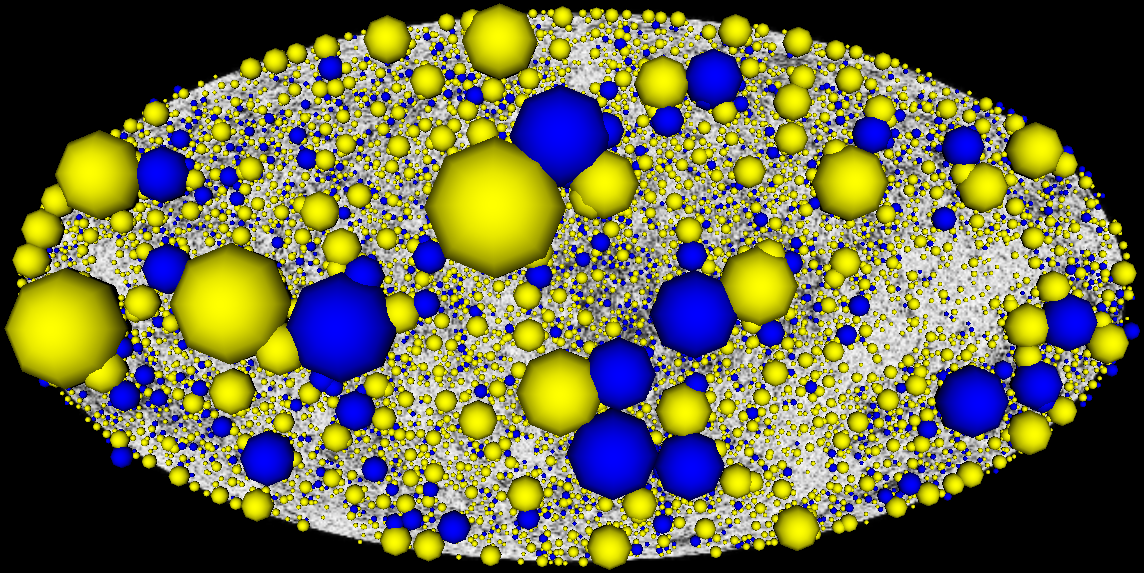}\\
    \includegraphics[width=0.5\textwidth]{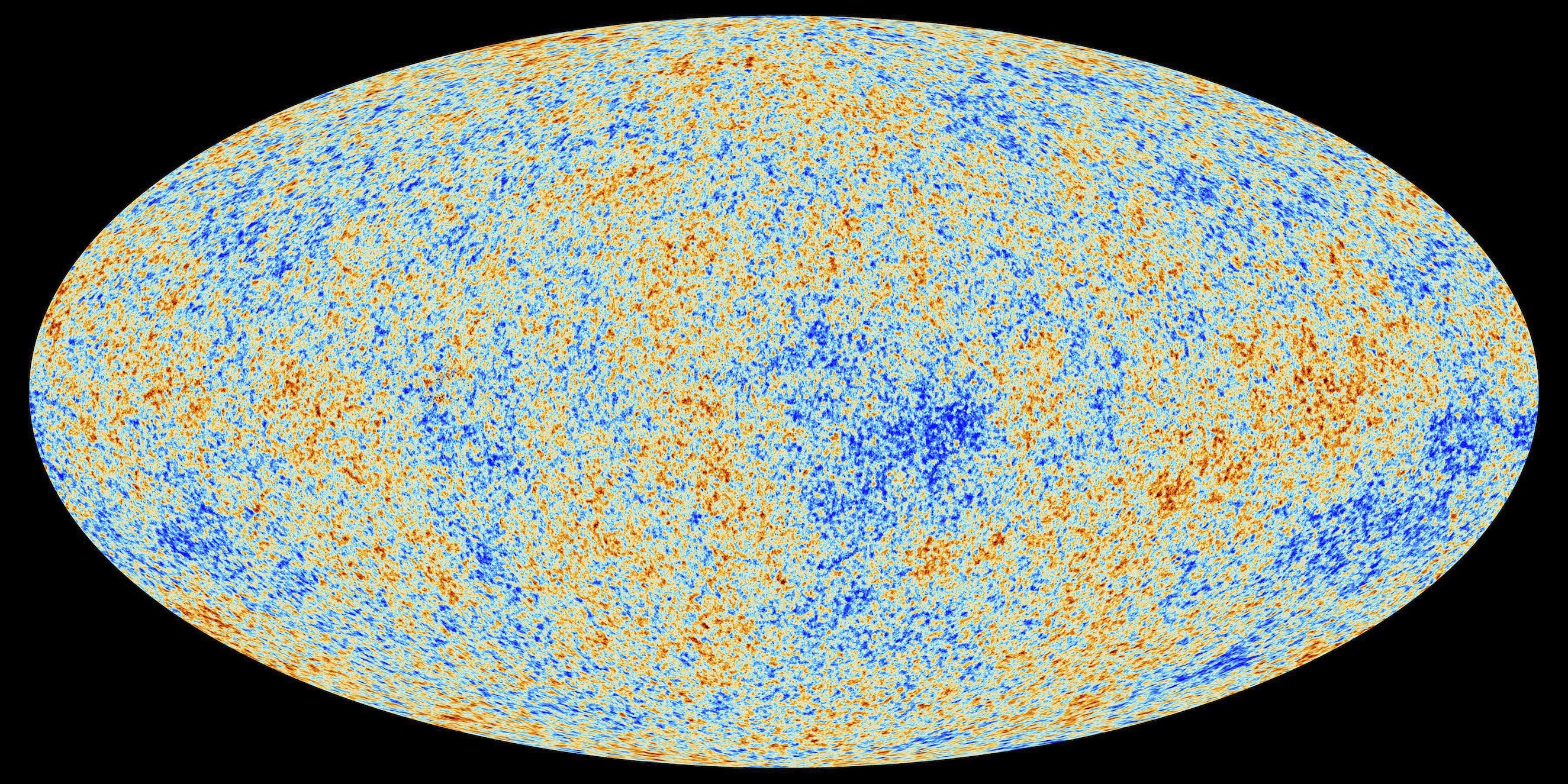}
    \caption{SIFT bubbles on a 2D cosmic microwave background map projection.}
    \label{fig:CMB-sift}
\end{figure}

Finally, in the case of a galactic simulation in Figure~\ref{fig:galaxy-sift}, image-domain features highlight concentrations and voids in the projected density: positive features (yellow) correspond to stars and other massive accumulations, negative features (blue) to interstellar spaces. A single large positive feature appears representative of the overall galactic mass concentration in the image; we do not claim that this explains galactic rotation-curve anomalies. Figure~\ref{fig:CMB-sift} shows features extracted on a 2D CMB map projection: note the bubble and droplet structure in the image domain, although geometry near the edges is distorted by the projection.

%\begin{figure}[h]
%    \centering
%    \includegraphics[width=0.45\textwidth]{figures/PrimatesThree.png}
%    \includegraphics[width=0.45\textwidth]{figures/PrimatesThreeGraph.png}
    %\caption{3D SIFT bubbles in the primate brain MRI. For these three subjects (macaque, chimpanzee, human), the number of features per brain is linearly related to the expected number of cortical neurons per species.}
%    \label{fig:macaque-sift}
%\end{figure}

%\begin{figure}[h]
%    \centering
    %\includegraphics[width=0.35\textwidth,angle=90]{MACAQUEMRI.png}
    %\includegraphics[width=0.35\textwidth,angle=90]{HUMANMRI.png}
    %\caption{3D SIFT bubbles in macaque and human brain MRI}
    %\label{fig:macaque-sift}
%\end{figure}

As can be seen, the bubbles emerge as features upon evolving the image through the Gaussian scale-space as a diffusion process and calculating peaks (and valleys) in the stack of scale-space images using difference-of-Gaussian (an efficient approximation of Laplacian-of-Gaussian). These extrema correspond to regions of density concentration or depletion that are identified at their characteristic scale in a manner invariant to the resolution of the image. We believe that the study of such features may help deep learning models better account for topology---for example, models challenged by knotted proteins and topologically complex folds~\cite{dabrowski2023alphafold}---though this remains speculative. Keeping this in mind, we are also exploring developing a diffusion-based deep learning model~\cite{sohl2015deep}, which schedules Gaussian noise of increasing variance similarly to Gaussian scale-space, where information concentrated in bubble-like features may inform the learning process. 

\subsection{Analogies to Chemistry and Cosmology}

In this section, we present analogies between the scale-invariant feature transform (SIFT), the free Schr\"odinger equation governing wavepacket spreading at the atomic and molecular scale, and Friedmann's equations governing universal expansion at the cosmological scale. These are mathematical and conceptual analogies based on shared Laplacian or scale-factor structure; they are not claims of physical identity.

\subsubsection*{Schr\"odinger's Equation}

%\section{Schrödinger's Equation and SIFT Atomic features}
%\subsection{An Atomical Analogy: Schrödinger's Equation}

Upon comparing the diffusion (heat), free time-dependent Schr\"odinger, and SIFT scale-space equations, we note a shared Laplacian-driven structure. The free Schr\"odinger equation
\[
i\hbar\,\frac{\partial \Psi}{\partial t}
= -\frac{\hbar^2}{2m}\,\nabla^2 \Psi
\]
governs how the wavefunction $\Psi(x,y,z,t)$ spreads, and its squared magnitude $|\Psi|^2$ is the probability density for finding a particle. Local maxima of an electron-density image (as in DFT) indicate regions of high electronic density, which for molecules typically coincide with nuclear positions. When we treat a molecular volume as an image and evolve it through Gaussian scale-space, the SIFT keypoint detector, which finds extrema of the scale-normalized Laplacian, identifies ``positive'' features at approximately those same atomic coordinates. We also observe ``negative'' features, which are located in topological gaps in the density.

\begin{figure}[h]
    \centering
    \includegraphics[width=0.5\textwidth]{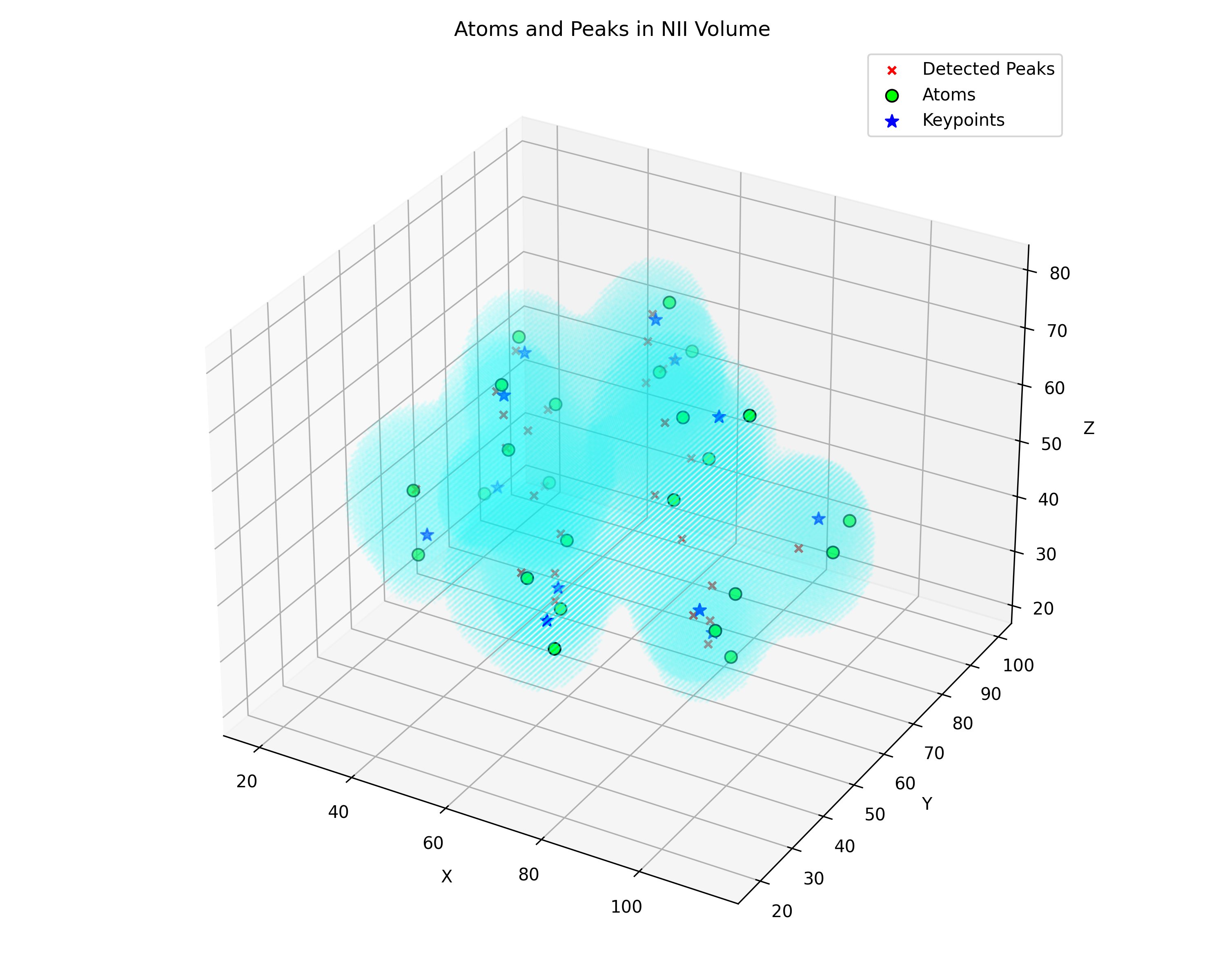}
    \caption{Atom peaks and feature alignment in fructose.}
    \label{fig:peaks}
\end{figure}

The close alignment between density peaks, atomic coordinates and SIFT keypoints is therefore expected for concentrated nuclear densities, rather than a derivation from molecular Schr\"odinger dynamics per se. As shown in Figure~\ref{fig:peaks}, the peaks detected as maxima in the DFT volume, the atomic positions specified in the molecule file, and the SIFT keypoint locations are in close alignment; the number of features matches the number of non-hydrogen atoms in this example.

Gaussian scale-space is related to free Schr\"odinger evolution by a Wick rotation, see Appendix~\ref{app:schrodinger_image}.

\subsubsection*{Hubble Expansion and Scale–Space Growth}

In cosmology, the Friedmann scale factor \(a(t)\) describes how spatial distances expand over cosmic time \(t\), and may be characterized in terms of the Hubble parameter $H(t)$ as:
\[
H(t) \;=\; \frac{1}{a(t)}\,\frac{da}{dt}.
\]
It measures the (relative) expansion rate of the universe, so that
\[
a(t) \;=\; a_0\exp\!\Bigl(\!\int_{t_0}^{t}H(\tau)\,d\tau\Bigr),
\]
where the initial scale factor at time \(t_0\) is \(a_0\). The evolution of the scale factor in the Friedmann--Lema\^\i tre--Robertson--Walker (FLRW) metric comes from solving the Friedmann equations.

We note that in the SIFT approach, an image evolves through scale-space according to the Gaussian parameter \(\sigma\), where \(\sigma = \sqrt{t}\) for diffusion time \(t\), reminiscent of a radiation-dominated universe where \(a(t)\propto \sqrt{t}\). This is a dimensional analogy only.
%It is also worth noting the Schwartzchild metrics, which is the other solution to the Einstien's field equations. Both diagonal metrics feature a scale‐factor‐like function (\(a(t)\) in FLRW, \(1 - 2GM/(c^2 r)\) in Schwarzschild) that governs distances, whether expanding–universe or gravitational–field contexts.

%This contrasts with de Sitter expansion but is still interesting from the point of view of how local image structure “evaporates” slower at coarse scales.

If we now write \(\sigma(t) = \sqrt{t}\), then by the chain rule:
\[
\frac{d\sigma}{dt} = \frac{1}{2\sqrt{t}} = \frac{1}{2\sigma},
\]
and so a natural analogue of the Hubble parameter in scale–space becomes:
\begin{equation}
H_{\sigma}(t) = \frac{1}{\sigma(t)} \cdot \frac{d\sigma}{dt} = \frac{1}{2\sigma^2}
\label{eq:hubble_scale}
\end{equation}

This rate is inversely proportional to \(\sigma^2\), meaning that the ``rate of expansion'' of scale diminishes as we go to higher scales. Now, SIFT uses a geometric progression for sampling the scale \(\sigma(t) = \sigma_t\), with a constant multiplicative factor \(k > 1\) such that
\[
\sigma_t = k\sigma_{t-1}.
\]
Scale growth over time is then
\[
\frac{d\sigma}{dt} \approx \frac{\sigma_t - \sigma_{t-1}}{1} = (k - 1)\sigma_{t-1}.
\]
Combining with \(\sigma_{t-1} = \sqrt{t - 1}\), one obtains a mnemonic relation
\begin{equation}
k - 1 \approx \frac{1}{\sigma}\,\frac{d\sigma}{dt} = \frac{1}{2\sigma^2},
\label{eq:hubble_SIFT}
\end{equation}
noting that $k-1$ is dimensionless while $1/(2\sigma^2)$ is not, and thus its correspondence to the Hubble parameter here is illustrative rather than dimensional.

Thus, Equations~\eqref{eq:hubble_scale} and~\eqref{eq:hubble_SIFT} present an analogy between SIFT scale sampling and a decaying Hubble-like expansion rate, with \(k - 1\) playing the role of an effective scale-space expansion increment, falling off with \(\sigma\). In practice, parameter $k = \sigma_t/\sigma_{t-1}$ is determined logarithmically by the number $N$ of samples per octave or doubling of scale $k^N=2$, where larger $N$ results in a higher sampling rate and precision, but larger memory and computational requirements. A number $N=3$ samples per octave (i.e.\ $k=2^{1/3}\approx1.26$) was found to maximize feature detection repeatability in natural images and thus balance sampling density and computational cost in the SIFT algorithm~\cite{lowe2004distinctive}.\footnote{We note the analogy to logarithmic sampling of frequency in musical tones, where the Western equal-tempered musical scale is defined by $N=12$ notes per octave and a frequency ratio $k=2^{1/12}\approx1.059$ between subsequent notes.}

% So while the analogy is not exact, it suggests how ideas from cosmology (like a scale factor and Hubble expansion) can help interpret Gaussian diffusion in image scale–space. The implications of this analogy are as follows.
% %Note also that the right side of Equation~\eqref{eq:hubble_ricci} represents half the Gaussian curvature of a sphere of radius $\sigma$ or equivalently the Ricci scalar linking mass to gravity in Einstein's field equations. 
% \begin{enumerate}
%   \item \textbf{Discrete “Expansion”:}  SIFT’s geometric progression of scales causes features to “grow” in scale–space by a fixed multiplicative factor each octave, much like exponential cosmic expansion.
%   \item \textbf{Feature Horizons:}  Just as objects beyond the cosmic horizon recede faster than light, image structures too small or too large may fall outside SIFT’s detectable scale range if the scale ratio is too large or small.
%   \item \textbf{Tuning the “Hubble Rate”:}  Choosing the scale increment \(\sigma_{k+1}/\sigma_k\) controls how densely scale–space is sampled, which helps optimizing for computational cost.
% \end{enumerate}

% Thus, treating \(\sigma\) as a “scale–factor” with constant relative growth rate \(H_s\) not only clarifies the octave structure of SIFT but also suggests deeper conceptual links, “feature horizons,” scale–space “cosmology,” and the balance between sampling density and computational cost.

\section{Discussion}

The black-box nature of neural networks~\cite{rudin2019stop} makes it increasingly relevant to understand the mechanical rules by which information transforms from one layer to the next. This work studies convolutional filtering~\cite{lecun2015deep} within the framework of relativistic energy--momentum, proposing a minimal theory based on real-valued linear filters, nonlinear rectified convolution and scalar image data. Computational demonstrations show Lorentz-like behaviour of image information at the pixel level, and emergent topological structure consistent with Morse theory across physical phenomena ranging from atoms to galaxies.

In the neural network setting, we place information mechanics in the context of the CNN architecture~\cite{lecun1989handwritten,fukushima1983neocognitron}, which ushered in modern AI following breakthrough accuracy in large-scale image recognition~\cite{krizhevsky2012imagenet}. Nonlinear rectification endows information with an analogue of relativistic momentum by rejecting negative correlation---an effect absent from classical linear image processing theories such as wavelets~\cite{haar1909theorie,gabor1946theory}. Rectified convolution was introduced relatively recently in artificial neural networks~\cite{nair2010rectified} and became widespread only after GPU-based training~\cite{krizhevsky2012imagenet}; its mechanical effect on image information was therefore not studied in those earlier frameworks. Recent work studies Lorentz equivariance explicitly in specialized networks and data~\cite{bogatskiy2020lorentz}; our focus instead is on generic filtering in image-based neural networks. Although we present the framework for scalar image filtering, it was inspired by recent work showing that primary sum $\Sigma$ and gradient $\nabla$ components of multi-channel generic CNN models account for over $92\%$ of accuracy in the settings of~\cite{frija2025mechanics}. Rectification is also common outside deep networks, for example in AC--DC power conversion and to enforce unidirectional reflection in graphics simulations.

In terms of physics, we propose an analogy between image evolution under invariant linear filters and quantum state evolution under Hermitian operators. The maximum speed of information in the image plane is limited by the discrete filter width to $dx/dt \le (Width-1)/2$ pixels per layer, analogous to the speed of light $c\approx3\times10^8$\,m\,s$^{-1}$ under the Lorentz transform. Filters decompose into symmetric and antisymmetric components that, under rectified convolution, function analogously to rest mass and momentum in relativistic energy: the symmetric component yields isotropic transformations that preserve the image centre of mass (e.g.\ intensity diffusion from the sum), while the antisymmetric component yields directional momentum-like displacement (e.g.\ vibrational and translational motion from gradient components). Note that while a typical quantum mechanical operator is represented as complex-valued square matrices, an information mechanical operator here refers to a real-valued vector.

Our model interprets CNN filtering in terms of the DCT spectrum, geometrical symmetry and differential operators, in a similar spirit to how quantum mechanics interprets physical phenomena through observable spectra, symmetry~\cite{heisenberg1926multibodyproblem} and differential operators~\cite{Schrodinger1926}. Using symmetric and antisymmetric filter components to demonstrate Lorentz-like motion and emergent Morse topology parallels Witten's demonstration that bosonic and fermionic supersymmetric quantum fluctuations lead to Morse topology and inequalities~\cite{witten1982supersymmetry}. Our minimal sum $\Sigma = [1,1]$ and difference $\nabla = [-1,1]$ operators are reminiscent of Winterberg's Planck Vacuum model~\cite{winterberg2003planck}---again a non-mainstream proposal---which uses finite average and difference operators to derive Maxwell's and Einstein's field equations as antisymmetric and symmetric modes of a vortex lattice. Whereas that model adopts a nonlinear Schr\"odinger equation on hypothetical positive and negative Planck mass particles, ours uses nonlinear rectified convolution on a field of non-negative pixels---an image superposition of binary information.

For experimental scientists using digital image processing, the Gaussian scale-space provides a unifying, scale-invariant description of structure across physical scales, including atom-scale electron density~\cite{popelier2000full}, human-scale MRI~\cite{toews2013efficient,chauvin2021efficient}, and galaxy-scale clusters and filaments~\cite{sousbie2011persistent}. Scale-space theory emerged from computer vision~\cite{koenderink1984structure,witkin1987scale,damon1995local}, with topological structure arising at scale parameter $\sigma \propto \sqrt{t}$ and with invariance to scale as well as Euclidean translation and rotation~\cite{batzner20223}. The scale-invariant (SIFT) feature~\cite{lowe2004distinctive} provides a common description of this structure according to Morse theory as imagined by Thom~\cite{thom1974stabilite} and Witten~\cite{witten1982supersymmetry}, consistent in spirit with Schr\"odinger evolution at the atomic scale and Friedmann's equations at the cosmological scale. Content around critical points may be encoded as invariant descriptors for indexing similar structure in large datasets in the context of memory-based learning~\cite{cover1967nearest}. For example, images of the same person or family members can be indexed reliably in large MRI collections~\cite{chauvin2021efficient}, and similar methods could potentially index biological or molecular motifs or inform architectures beyond local CNN correlations, such as transformers.

Limitations remain. The relativistic motion we demonstrate follows a linear Lorentz-transform curve closely but not exactly: displacements in Figure~\ref{fig:prop_graph} match expectation at $\beta = 0$ and $\beta = 1$, but are slightly slower or faster for intermediate $\beta \in (0, 1)$. We focused on primary sum and gradient components; other DCT modes, such as the antisymmetric saddle $(\ell, k)=(1,1)$ or symmetric Laplacian $(\ell, k)=(2,2)$, merit further study. The DCT gradient achieves maximum propagation displacement $(Width - 1)/2$ only for discrete $2\times2$ and $3\times3$ filters, not larger ones. Finally, square filters coarsely approximate ideal round or spherical filters, and diagonal distances scale by $\sqrt{2}$ relative to horizontal and vertical ones.

\section{Conclusions}
\label{conclusions}

This work studies convolutional neural network filtering within the framework of relativistic energy--momentum. A filter may be represented as the sum of radially symmetric and antisymmetric components that, under rectified convolution, act analogously to rest mass and momentum. Information mechanics thus offers a simple analogy between CNN filtering and quantum state evolution, with Lorentz-like motion in the image plane arising from the proportion of antisymmetric filter energy---proportional to momentum in the Lorentz transform.

Repeated filtering leads to the Gaussian scale space and a topological description from scale-invariant Morse critical points. Although these have long been known, we present the structure of physical phenomena from atomic to galactic scales, demonstrating how critical points identify either maximally spherical droplet-like density concentrations (atoms, stars) and bubble-like regions of density depletion between them. We hope these visualizations convey an intuition for stable yet sometimes invisible features of physical structure, much as Faraday's iron-filing demonstrations made magnetic field lines visible.

Future work will extend the theory to modern architectures such as attention-based models~\cite{vaswani2017attention} and diffusion models~\cite{sohl2015deep}, under the unifying umbrella of information mechanics, and further study diffusion bubble structure across molecular, cellular, biological and cosmological contexts.

\section{Acknowledgements}
The mechanics of rectified convolution was first studied in the masters' work of Liam Frija-Altarac~\cite{frija2025mechanics}. This work was supported by the Canadian Natural Sciences and Engineering Research Council (NSERC) Discovery Grant of Matthew Toews.

\section{Appendices}
\appendix
\section{$3\times3$ Image Decomposition into Radially Symmetric and Antisymmetric Components }
\label{app:decomp-example}

Consider decomposing an example $3\times3$ image
\[
I = \begin{bmatrix}
7 & 8 & 9\\
6 & 1 & 2\\
5 & 4 & 3
\end{bmatrix}.
\]
With centre at \((1,1)\), the three radial shells are:
\begin{itemize}
  \item \(r=0\): \(\{(1,1)\}\) with value \(1\)
  \item \(r=1\): \(\{(1,2),(2,1),(0,1),(1,0)\}\) with values \(\{2,4,6,8\}\); average
    \[
      \frac{2+4+6+8}{4} = 5
    \]
  \item \(r=\sqrt2\): \(\{(2,2),(2,0),(0,0),(0,2)\}\) with values \(\{3,5,7,9\}\); average
    \[
      \frac{3+5+7+9}{4} = 6
    \]
\end{itemize}

Hence the radially symmetric component is
\[
I_s = \begin{bmatrix}
6 & 5 & 6\\
5 & 1  & 5\\
6 & 5 & 6
\end{bmatrix}
\]
and the antisymmetric (shell-demeaned) remainder
\[
I_a = I - I_s
= \begin{bmatrix}
7-6 & 8-5 & 9-6\\
6-5 & 1-1 & 2-5\\
5-6 & 4-5 & 3-6
\end{bmatrix}
\]
\[= \begin{bmatrix}
1 & 3 & 3\\
1 &  0   &  -3\\
 -1 &  -1 &  -3
\end{bmatrix}.\]

Finally, to check orthogonality, compute the inner product
\[
\langle I_s, I_a\rangle
= \sum_{i,j} I_s(i,j)\,I_a(i,j).
\]
The four corner terms sum to zero:
\[
6\bigl(1+3-1-3\bigr)=0,
\]
and the four edge terms likewise cancel:
\[
5\bigl(3+1-3-1\bigr)=0,
\]
while the centre contributes \(1\cdot0=0\).
Therefore \(\langle I_s,I_a\rangle=0\), confirming orthogonality.

\section{Diffusion and Heat Equation}
\label{app:Diffusion}

\subsubsection*{Governing PDE and Conditions}

Let
\[
u(x,y,t)\;\colon\;[0,a]\times[0,b]\times[0,\infty)\;\to\;\mathbb{R}
\]
denote the temperature at point \((x,y)\) of a rectangular plate at time \(t\ge0\).  We assume:
\begin{itemize}
  \item \textbf{Heat equation:}
    \begin{equation}\label{heat-pde}
      \frac{\partial u}{\partial t} \;=\; \kappa^2\,\bigl(u_{xx} + u_{yy}\bigr),
    \end{equation}
    where \(\kappa^2>0\) is the thermal diffusivity (units: \(\text{length}^2/\text{time}\)), and subscripts denote partial derivatives.  
  \item \textbf{Dirichlet boundary conditions:}
    \[
      u(0,y,t)=u(a,y,t)=u(x,0,t)=u(x,b,t)=0
    \]
    \[t\ge0\]
  \item \textbf{Initial condition:}
    \[
      u(x,y,0)=f(x,y),
      \quad (x,y)\in[0,a]\times[0,b]
    \]
    where \(f(x,y)\) is the prescribed initial temperature distribution.
\end{itemize}

\subsubsection*{Separation of Variables}

Assume a product solution of the form
\[
u(x,y,t)=X(x)\,Y(y)\,T(t)
\]
Substituting into \eqref{heat-pde} gives
\[
X(x)\,Y(y)\,T'(t)
\]
\[= \kappa^2\,\bigl[X''(x)\,Y(y) + X(x)\,Y''(y)\bigr]\,T(t)\]
Divide both sides by \(\kappa^2\,X(x)\,Y(y)\,T(t)\) to obtain
\[
\frac{1}{\kappa^2}\,\frac{T'(t)}{T(t)}
= \frac{X''(x)}{X(x)} + \frac{Y''(y)}{Y(y)}
= -\lambda
\]
where \(\lambda\) is a separation constant.  Hence we get two spatial ODEs and one temporal ODE:

\begin{align}
  T'(t) + \lambda\,\kappa^2\,T(t) &= 0
  \label{heat-time}\\
  X''(x) + \mu^2\,X(x) &= 0\\
  Y''(y) + \nu^2\,Y(y) = 0
  \label{heat-space}
\end{align}
with \(\lambda = \mu^2 + \nu^2\), where \(\mu\) and \(\nu\) will be determined by boundary conditions.

\subsubsection*{Spatial Eigenvalue Problems}

Impose Dirichlet conditions on \(X\) and \(Y\):
\[
X(0)=X(a)=0,\quad Y(0)=Y(b)=0
\]
The standard eigenvalues are
\[
\mu_m = \frac{m\pi}{a},\quad m=1,2,\ldots
\]
\[\nu_n = \frac{n\pi}{b},\quad n=1,2,\ldots\]
with the corresponding eigenfunctions
\[
X_m(x) = \sqrt{\frac{2}{a}}\;\sin\!\Bigl(\frac{m\pi x}{a}\Bigr)
\]
\[Y_n(y) = \sqrt{\frac{2}{b}}\;\sin\!\Bigl(\frac{n\pi y}{b}\Bigr)\]
and the corresponding separation constant is
\[
\lambda_{mn} \;=\; \mu_m^2 + \nu_n^2 \;=\; \Bigl(\tfrac{m\pi}{a}\Bigr)^2 + \Bigl(\tfrac{n\pi}{b}\Bigr)^2
\]

\subsubsection*{Temporal ODE and Solution}

From \eqref{heat-time}, for each pair \((m,n)\),
\[
T' + \kappa^2\,\lambda_{mn}\,T = 0
\Longrightarrow
T_{mn}(t) = \exp\bigl(-\,\kappa^2\,\lambda_{mn}\,t\bigr)
\]
Thus each mode decays exponentially in time with rate \(\kappa^2\,\lambda_{mn}\).

\subsubsection*{General Solution}

Superposing over all integer modes \(m,n\) and matching the initial condition via coefficients \(A_{mn}\), the solution is
\[
u(x,y,t)=\]
\[\sum_{m=1}^\infty\sum_{n=1}^\infty
A_{mn}\,
X_m(x)\,Y_n(y)\,
\exp\bigl(-\,\kappa^2(\mu_m^2+\nu_n^2)\,t\bigr)
\]
where
\[
A_{mn}
= \int_{0}^{a}\int_{0}^{b}
f(x,y)\,X_m(x)\,Y_n(y)\,dx\,dy
\]
and \(X_m, Y_n\) are as above.

\subsubsection*{Unbounded‐Domain Solution and Connection to SIFT}

In many imaging applications, particularly SIFT, we treat the image as defined on the entire plane \(\mathbb{R}^2\).  Hence, the heat equation becomes:
\[
\frac{\partial u}{\partial t}
= \kappa^2\bigl(u_{xx}+u_{yy}\bigr),
\quad
u(x,y,0)=f(x,y)
\]
\[(x,y)\in\mathbb{R}^2\]
and its fundamental solution (heat kernel) is
\[u(x,y,t)\]
\begin{equation}\label{eq:heat-unbounded}
= \iint_{\mathbb{R}^2}
G(x-\xi,y-\eta;\,t)\;
f(\xi,\eta)\;d\xi\,d\eta
\end{equation}
\[
G(x,y;\,t)
= \frac{1}{4\pi\,\kappa^2\,t}
\exp\!\Bigl(-\frac{x^2+y^2}{4\,\kappa^2\,t}\Bigr)
\]
Equivalently, one writes
\[
u(\,\cdot\,,t)
= G(\,\cdot\,;\,t)\ast f
\quad\text{on }\mathbb{R}^2
\]

\subsection*{Scale–Space in SIFT}

In the SIFT framework, we identify
\[
I(x,y,\sigma)\;\longleftrightarrow\;u(x,y,t)
\]
subject to the standard \emph{scale–time} relation
\[
t \;=\;\sigma^2
\]
so that Gaussian smoothing at scale \(\sigma\) is exactly the diffusion solution at time \(t=\sigma^2\).  We denote the diffusion constant in image space as \(D>0\), so that
\[
\frac{\partial I}{\partial \sigma}
= 2D\sigma\,\bigl(I_{xx}+I_{yy}\bigr),
\quad
I(x,y,0)=I_0(x,y)
\]
Keypoints are detected as extrema of the Difference–of–Gaussians,
\[
\DoG(x,y;\sigma)
= I(x,y;\,k\sigma)\;-\;I(x,y;\,\sigma)
\]
which, up to a constant factor, approximates the Laplacian:
\[
\DoG\;\propto\;\sigma^2\,\nabla^2 I
\]
By modeling \(I\) on \(\mathbb{R}^2\), boundary artifacts are avoided for moderate \(\sigma\), since real images are typically much larger than the effective support of the Gaussian kernel.
\section{Scale–Space Schrödinger Analogy}
\label{app:schrodinger_image}

We now recast the 2D time–dependent Schrödinger equation into the language of image diffusion through scale–space. Recall the free–particle Schrödinger PDE on \(\mathbb{R}^2\):
\[
i\,\hbar\,\frac{\partial \Psi}{\partial t}
= -\,\frac{\hbar^2}{2m}\,\nabla^2 \Psi,
\quad
\Psi(x,y,0)=\Psi_0(x,y)
\]

To draw the analogy with diffusion in image space, we perform a Wick rotation:
\[
t \mapsto i\sigma^2
\]
which transforms unitary time evolution into real diffusion. Under this substitution, the time derivative becomes:
\[
\frac{\partial \Psi}{\partial t}
= \frac{1}{dt/d\sigma} \cdot \frac{\partial \Psi}{\partial \sigma}
= \frac{1}{2i\sigma} \cdot \frac{\partial \Psi}{\partial \sigma}
\]

Substituting into Schrödinger's equation yields:
\[
i\hbar \left( \frac{1}{2i\sigma} \frac{\partial \Psi}{\partial \sigma} \right)
= -\,\frac{\hbar^2}{2m} \nabla^2 \Psi
\]
\[\Rightarrow\quad
\frac{\hbar}{2\sigma} \frac{\partial \Psi}{\partial \sigma}
= \frac{\hbar^2}{2m} \nabla^2 \Psi\]

Now define the image:
\[
I(x,y,\sigma) \coloneqq \Psi(x,y,\,i\sigma^2)
\]
which satisfies the PDE:
\[
\frac{\partial I}{\partial \sigma}
= 2\sigma D \nabla^2 I,
\quad \text{where } D = \frac{\hbar}{2m}
\]

This resembles the form of the scale–space diffusion equation, where \(\sigma\) plays the role of the square root of diffusion time $\sigma = \sqrt{t}$.

\subsection*{Separation of Variables}

Assume
\[
I(x,y,\sigma)=\Phi(x,y)\;T(\sigma)
\]
Plug into the PDE:
\[
\Phi\,T'
= 2\,D\,\sigma\;(\nabla^2\Phi)\,T
\]
Divide by \(\Phi\,T\) and set equal to \(-\lambda\):
\[
\frac{T'}{T}
= \frac{2\,D\,\sigma}{\Phi}\,\nabla^2\Phi
= -\lambda
\]
This yields the pair
\begin{align}
&T' + \lambda\,T = 0
\label{diff-time}\\[6pt]
&\nabla^2\Phi + \frac{\lambda}{2\,D\,\sigma}\,\Phi = 0
\label{diff-space}
\end{align}

\subsection*{Solutions}

Scale–factor equation \eqref{diff-time}
\[
T(\sigma) = T(0)\,\exp(-\lambda\,\sigma)
\]

Consider a spatial eigenproblem where \eqref{diff-space} \(k_x, k_y\) are spatial frequencies along x and y spatial dimensions, respectively. In the context of image space, these would indicate how "frequently" pixel values change while travelling along the x and y dimensions of an image.

On \(\mathbb R^2\) take \[\Phi_{k}(x,y)=e^{\,i(k_xx+k_yy)}\]
\[\nabla^2\Phi=-|k|^2\Phi\]

where, \(|k| = \sqrt{k_x^2 + k_y^2}\).

Then
\(\lambda = 2\,D\,\sigma\,|k|^2\)

\paragraph{Combined mode}  
Substitute \(\lambda\) back into \(T\) and multiply by \(\Phi\):
\[
I_{k}(x,y,\sigma)
= A_k\,
e^{\,i(k_xx+k_yy)}\,
e^{-2\,D\,\sigma\,|k|^2\;\sigma}
\]
\[= A_k\,e^{\,i(k\!\cdot\!x)}\,e^{-D\,|k|^2\,\sigma^2}\]
which is exactly the Gaussian‐scale‐space blur of variance \(\sigma^2\).

\bigskip
\textbf{Note:} the two factors of \(\sigma\) ensure that under \(t=\sigma^2\) we recover the usual heat‐kernel variance.

\paragraph{Summary:}
By Wick-rotating the Schrödinger equation via \(t = i\sigma^2\), unitary wave evolution becomes Gaussian diffusion. The wavefunction \(\Psi(x,y,t)\) maps to an evolving image \(I(x,y,\sigma)\), and the propagator becomes the familiar Gaussian kernel used in SIFT and other scale–space methods.

\section{Derivation of the Lorentz Transformation}
\label{app:lorentz}

Starting with the Lorentz transformation:

Let
\[
T: S \to S'
\]
be a transformation of frames of reference, where
\[
S = \begin{bmatrix}x\\ct\end{bmatrix},\quad
S' = \begin{bmatrix}x'\\ct'\end{bmatrix}
\]
\begin{equation}
T\begin{bmatrix}x\\ct\end{bmatrix}
= \begin{bmatrix}x'\\ct'\end{bmatrix}
\end{equation}

\subsection{Assumptions}
\begin{enumerate}
  \item $T$ is linear.
  \item The speed of light $c$ is constant in all inertial frames.
  \item $T\begin{bmatrix}\beta\\1\end{bmatrix}
        = \mu\begin{bmatrix}0\\1\end{bmatrix}$,
        where $\beta=v/c$.
  \item $T_v T_{-v} = I$, i.e.\ reversing the boost returns the identity.
\end{enumerate}

\subsection{Background on Assumption 3}
A boost with velocity $v$ maps the worldline of the moving origin. Writing $\beta=v/c$, the condition
\[
T\begin{bmatrix}\beta\\1\end{bmatrix}
= \mu\begin{bmatrix}0\\1\end{bmatrix}
\]
encodes that this worldline is mapped to the time axis of the primed frame (up to a scale factor $\mu$).

\subsection{Background on Assumption 4}
A second boost by $-v$ must undo the first, so
\[
T_v T_{-v} = I
\]

\subsection{Derivation of the Matrix Form}
We seek a $2\times2$ matrix $T$ diagonalizable with eigenvectors
$\bigl[1,1\bigr]^T$, $\bigl[-1,1\bigr]^T$.  Imposing 
\[
T\begin{bmatrix}1\\1\end{bmatrix}
= \lambda_+\begin{bmatrix}1\\1\end{bmatrix},\quad
T\begin{bmatrix}-1\\1\end{bmatrix}
= \lambda_-\begin{bmatrix}-1\\1\end{bmatrix}
\]
and using Assumption~3 leads to the standard form
\[
T = \gamma
\begin{bmatrix}
1 & -\beta\\
-\beta & 1
\end{bmatrix},\quad
\gamma=\frac1{\sqrt{1-\beta^2}}
\]

\section{Four-Vector Representation}

Also known as $\mu$ (mu) notation. We use the mostly-minus Minkowski metric convention $(+,-,-,-)$.

\[
\chi^\mu = (ct, x, y, z), \quad \text{with}\quad \mu = 0, 1, 2, 3,
\]
\[
\chi_\mu = g_{\mu\nu}\chi^\nu = (ct, -x, -y, -z).
\]

\begin{align*}
    \chi^\mu &: \text{contravariant components of the four-vector}, \\
    \chi_\mu &: \text{covariant components of the four-vector}.
\end{align*}

\[
    \chi^\mu \, \chi_\mu =  c^2t^2 - x^2 - y^2 - z^2
\]
(summation over $\mu$ implied). More generally, for two different four-vectors $A^\mu$ and $B^\mu$,
\[
    A^\mu B_\mu = A^0 B_0 + A^1 B_1 + A^2 B_2 + A^3 B_3,
\]
using the Einstein summation convention.

Contravariant and covariant vectors are related by the metric tensor \(g_{\mu\nu}\),
\[
g_{\mu\nu} = \begin{bmatrix}
    1 & 0 & 0 & 0 \\
    0 & -1 & 0 & 0 \\
    0 & 0 & -1 & 0 \\
    0 & 0 & 0 & -1
\end{bmatrix}.
\]

The above metric is the Minkowski metric in the \((+, -, -, -)\) signature; an equivalent \((-, +, +, +)\) representation is also common. The tensors \(g_{\mu\nu}\) and \(g^{\mu\nu}\) are inverses of each other, and
\begin{align*}
    \chi_\mu &= g_{\mu\nu}\chi^\nu, \\
    \chi^\mu &= g^{\mu\nu}\chi_\nu.
\end{align*}

\section{Dirac's Equation}
\label{app:dirac}

Three principles of Dirac's equation:

\begin{enumerate}
    \item \( P^\mu \, P_\mu = (mc)^2 \rightarrow \left(\frac{E}{c}\right)^2 - \overline{P}^2 = (mc)^2 \)
    
    \item \( E \) \& \( P \) are quantum mechanical operators:
    \begin{align*}
        P_\mu &= i\hbar \partial_\mu
    \end{align*}
    $P_\mu$ and $\partial_\mu$ are four-momentum and four-gradient operators, respectively.
    Splitting \( P_\mu \) yields \(\hat{E}\) and \(\hat{P}\):
    \begin{align*}
        \hat{E} &= i\hbar \frac{\partial}{\partial t}, \\
        \hat{P} &= -i\hbar \left[ \frac{\partial}{\partial x}, \frac{\partial}{\partial y}, \frac{\partial}{\partial z} \right].
    \end{align*}

    \item First-order in time and space
\end{enumerate}

\subsection{Background on principle 1}

\[
P^\mu P_\mu = P^\mu g_{\mu\nu} P^\nu
\]
\[
= g_{00}(P^0)^2 + g_{11}(P^1)^2 + g_{22}(P^2)^2 + g_{33}(P^3)^2
\]

using the Minkowski metric \((+, -, -, -)\) notation

\[(P^0)^2-((P^1)^2+(P^2)^2+(P^3)^2)\]

\[= \left(\frac{E}{c}\right)^2 - ((P_x)^2 + (P_y)^2 + (P_z)^2)\]

\[= \left( \frac{E}{c}\right)^2 - \left|\overline{P}\right|^2\]

Hence, 
\begin{equation}
    P^\mu P_\mu = \left( \frac{E}{c}\right)^2 - \left|\overline{P}\right|^2.
\end{equation}

Also,
\begin{equation}
    E^2 = (mc^2)^2 +(pc)^2,
\end{equation}
so
\[P^\mu P_\mu = (mc)^2.\]

Now,
\begin{equation}
\left(\frac{E}{c}\right)^2 - P_x^2 - P_y^2 - P_z^2 = (mc)^2.
\end{equation}

\textbf{Goal:} Make \(E\) and \(P\) first-order derivatives. \\
One way to do this is to take square roots on both sides of Equation (D3):

\[\sqrt{\left(\frac{E}{c}\right)^2 - P_x^2 - P_y^2 - P_z^2} = mc.\]

But how do we factor out the squares? The answer is to find a way to make the square root linear. A formulation of this problem can look like this:

\[A\left(\frac{E}{c}\right) + BP_x + CP_y + DP_z
= \sqrt{\left(\frac{E}{c}\right)^2 - P_x^2 - P_y^2 - P_z^2}.\]

Square both sides:

\[\left(A\left(\frac{E}{c}\right) + BP_x + CP_y + DP_z\right)^2
= \left(\frac{E}{c}\right)^2 - P_x^2 - P_y^2 - P_z^2.\]

On comparing both sides, after expanding the LHS:

\[A^2 = 1, \quad B^2 = 1, \quad C^2 = 1, \quad D^2 = 1,\]
\[AB + BA = 0, \quad \ldots \quad AD + DA = 0.\]

\(AB + BA = 0\) implies that \(A\) and \(B\) do not commute, so they cannot be complex (or real) numbers; one must look to other domains, for example matrices. Dirac solved this problem and found four $4\times4$ matrices. The solution proposed by Dirac looks like this:

\[\gamma^0, \gamma^1, \gamma^2, \gamma^3  \rightarrow \text{called Dirac/gamma matrices}\]

The specific form of the $4\times4$ gamma matrices is non-unique, in the standard Dirac equation they are as follows. Matrix $\gamma^0$ is associated with the partial time derivative $\partial_0=\frac{\partial}{\partial_t}$, along with the $4\times4$ identity matrix $I_4$ associated with rest mass $m$ forms the diagonal component of the Dirac equation.

\[
I_4 = \begin{pmatrix}
1 & 0 & 0 & 0 \\
0 & 1 & 0 & 0 \\
0 & 0 & 1 & 0 \\
0 & 0 & 0 & 1
\end{pmatrix}
\quad
\gamma^0 = \begin{pmatrix}
1 & 0 & 0 & 0 \\
0 & 1 & 0 & 0 \\
0 & 0 & -1 & 0 \\
0 & 0 & 0 & -1
\end{pmatrix}
\]

The three $4\times4$ skew-diagonal gamma matrices may be written as
\[
\gamma^i = \begin{pmatrix}
0 & \sigma^i \\
-\sigma^i & 0
\end{pmatrix}, \quad i=1,2,3
\]

where $\sigma^1,\sigma^2,\sigma^3$ are the Pauli matrices 

\[
\sigma^1 = \begin{pmatrix} 0 & 1 \\ 1 & 0 \end{pmatrix}, \quad
\sigma^2 = \begin{pmatrix} 0 & -i \\ i & 0 \end{pmatrix}, \quad
\sigma^3 = \begin{pmatrix} 1 & 0 \\ 0 & -1 \end{pmatrix}
\]

associated with the partial spatial derivatives $\frac{\partial}{\partial_x},\frac{\partial}{\partial_y},\frac{\partial}{\partial_z}$ in the $x,y,z$ directions. Explicitly, these are:

\[
\gamma^1 = \begin{pmatrix}
0 & 0 & 0 & 1 \\
0 & 0 & 1 & 0 \\
0 & -1 & 0 & 0 \\
-1 & 0 & 0 & 0
\end{pmatrix},
\quad
\gamma^2 = \begin{pmatrix}
0 & 0 & 0 & -i \\
0 & 0 & i & 0 \\
0 & i & 0 & 0 \\
-i & 0 & 0 & 0
\end{pmatrix},
\]
\[\gamma^3 = \begin{pmatrix}
0 & 0 & 1 & 0 \\
0 & 0 & 0 & -1 \\
-1 & 0 & 0 & 0 \\
0 & 1 & 0 & 0
\end{pmatrix}\]

\[A = \gamma^0, \quad B =\gamma^1, \quad C=\gamma^2, \quad D = \gamma^3\]

also, \[(\gamma^0)^2 = 1, \quad (\gamma^1)^2 = (\gamma^2)^2 = (\gamma^3)^2 = -1 \]

and gamma matrices are anti commutative, \[\gamma^\mu \gamma^v + \gamma^v \gamma^\mu = 0 \quad \quad \quad \mu \neq v\]

Now, \[A\left(\frac{E}{c}\right) + BP_x + cP_y + DP_z\ \]
\[= \sqrt{\left(\frac{E}{c}\right)^2 - P_x^2 - P_y^2 - P_z^2}\]

\[\gamma^0\left(\frac{E}{c}\right) + \gamma^1P_x + \gamma^2P_y + \gamma^3P_z\ \]

\[=\sqrt{\left(\frac{E}{c}\right)^2 - P_x^2 - P_y^2 - P_z^2} = mc\]

also written as, \[\gamma^\mu P_\mu = mc\] using Einstein's summation convention.

Also, if we replace the four-vectors, we get: \[\gamma^ \mu(i \hbar \partial \mu) = mc \quad \rightarrow \quad [ i \hbar\gamma^\mu \partial \mu - mc] = 0\]

This is called Dirac's operator:
\[\mathcal{D} = \left[ i \hbar \gamma^\mu \partial \mu - mc\right]\]

It can be applied on the wave function as \(\mathcal{D}\psi\), where \(\psi\) is the four-vector form of the wave function, as follows:

% \[
% \left[ i\hbar \left( 
% \begin{pmatrix} 1 & 0 & 0 & 0 \\ 0 & 1 & 0 & 0 \\ 0 & 0 & -1 & 0 \\ 0 & 0 & 0 & -1 \end{pmatrix} \frac{\partial}{\partial x^0} + 
% \begin{pmatrix} 0 & 0 & 0 & 1 \\ 0 & 0 & 1 & 0 \\ 0 & -1 & 0 & 0 \\ -1 & 0 & 0 & 0 \end{pmatrix} \frac{\partial}{\partial x^1} + 
% \begin{pmatrix} 0 & 0 & 0 & -i \\ 0 & 0 & i & 0 \\ 0 & i & 0 & 0 \\ -i & 0 & 0 & 0 \end{pmatrix} \frac{\partial}{\partial x^2} + 
% \begin{pmatrix} 0 & 0 & 1 & 0 \\ 0 & 0 & 0 & -1 \\ -1 & 0 & 0 & 0 \\ 0 & 1 & 0 & 0 \end{pmatrix} \frac{\partial}{\partial x^3} 
% \right) - mc 
% \begin{pmatrix} 1 & 0 & 0 & 0 \\ 0 & 1 & 0 & 0 \\ 0 & 0 & 1 & 0 \\ 0 & 0 & 0 & 1 \end{pmatrix} 
% \right] \psi = 0 \]

\[
\left[ i\hbar \left( 
\gamma^0 \frac{\partial}{\partial x^0} + 
\gamma^1 \frac{\partial}{\partial x^1} + 
\gamma^2 \frac{\partial}{\partial x^2} + 
\gamma^3\frac{\partial}{\partial x^3} 
\right) - I_4mc
\right] \psi\] 
\[ \\ = 0 \]

\section{Experimental code and data}
\label{app:sift_imgs}

The C++ source code for 3D SIFT feature extraction may be found at \url{https://github.com/3dsift-rank}

Compiled binary files for 3D feature extraction (Windows, Linux, Mac) may be found at \url{www.matthewtoews.com}
    
Scrips for experiments extracting SIFT features and make diffusion bubbles can be found at: \url{https://github.com/aryanshukla7/Diffusion-Bubbles.git}. 

Code for generating information propagation results may be found at \url{https://github.com/liamaltarac/Information-Mechanics}.

Visualizations 3D bubbles were generated using the 3D Slicer software:
\url{https://www.slicer.org/}

The macaque brain MRI was acquired at 0.5mm voxel size on a 3T Siemens Prisma scanner, described in the following reference:
Valcourt Caron, Alex, et al. "In vivo submillimeter diffusion MRI dataset of 9 macaque brains curated for tractography." Scientific Data 12.1 (2025): 1199.

%The macaque brain MRI was acquired at 0.664 mm voxel size on a 1.5-Tesla Siemens Sonata, and provided by the Montreal Neurological Institute:
%\url{https://www.bic.mni.mcgill.ca/ServicesAtlases/MacaqueDownload}
%Stephen Frey, Deepak N. Pandya, M. Mallar Chakravarty, Lara Bailey, Michael Petrides, D. Louis Collins. ‘An MRI based average macaque monkey stereotaxic atlas and space (MNI monkey space)’, NeuroImage (2011).

The chimpanzee brain MRI was acquired at 0.6 mm voxel size in a 3-Tesla Siemens Trio MRI, provided National Chimpanzee Brain Resource (NS092988) at:
\url{https://www.chimpanzeebrain.org/mri-datasets-for-direct-download}

The human brain MRI was acquired at 0.7 mm voxel size on a 3-Tesla Siemens Skyra “Connectom” scanner, and was provided by the Human Connectome Project, WU-Minn Consortium (Principal Investigators: David Van Essen and Kamil Ugurbil; 1U54MH091657) funded by the 16 NIH Institutes and Centers that support the NIH Blueprint for Neuroscience Research; and by the McDonnell Center for Systems Neuroscience at Washington University. MRI \url{https://www.humanconnectome.org/}

Galactic simulation source code: \url{http://beltoforion.de/galaxy/galaxy_en.html}

Electron density images for molecules are generated based on density-functional theory (DFT) using code: \url{https://gitlab.com/gpaw/gpaw} with atomic arrangements provided in structured data file (SDF) format from:
\url{https://pubchem.ncbi.nlm.nih.gov/}. The silicon crystal was provided by Karim Zongo.

Lung CT images provided by the COPDGene project~\cite{regan2011genetic}.

3D video game scene (Minecraft) provided by Aryan Shukla, 2D display by Edouard Toews.

\begin{figure}[h]
    \centering
    \includegraphics[width=0.46\textwidth]{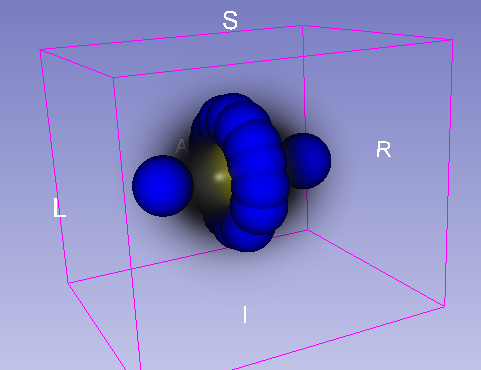}
    \caption{3D SIFT bubbles from a $H_2$ molecule}
    \label{fig:h2o-sift-1}
\end{figure}

\begin{figure}[h]
    \centering
    \includegraphics[width=0.46\textwidth]{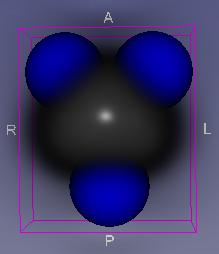}
    \caption{3D SIFT bubbles from a $H_2 O$ molecule}
    \label{fig:h2o-sift-2}
\end{figure}

\begin{figure}[h]
    \centering
    \includegraphics[width=0.46\textwidth]{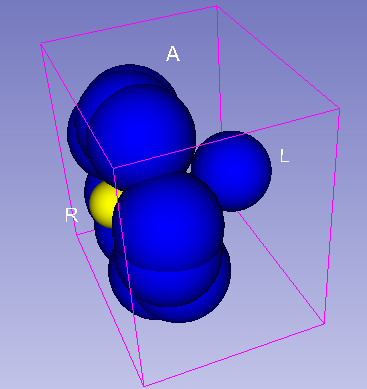}
    \caption{3D SIFT bubbles from a $HF$ molecule}
    \label{fig:hf-sift-1}
\end{figure}

\begin{figure}[h]
    \centering
    \includegraphics[width=0.46\textwidth]{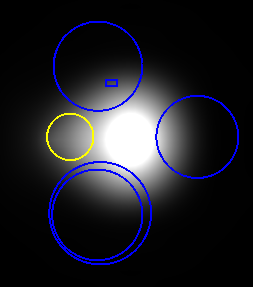}
    \caption{3D SIFT bubbles from a $HF$ molecule (2D slice)}
    \label{fig:hf-sift-2}
\end{figure}

\begin{figure}[h]
    \centering
    \includegraphics[width=0.46\textwidth]{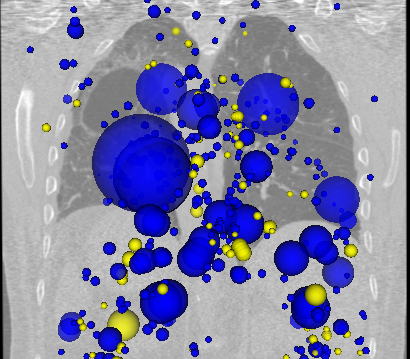}
    \caption{3D SIFT on Lung Images affected by chronic obstructive pulmonary disease (COPD).}
    \label{fig:lung-sift-1}
\end{figure}

\begin{figure}[h]
    \centering
    \includegraphics[width=0.46\textwidth]{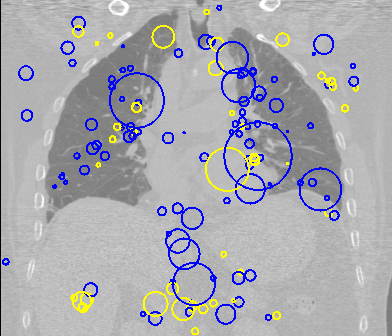}
    \caption{3D SIFT on Lung Images (2D coronal slice)}
    \label{fig:lung-sift-2}
\end{figure}

\begin{figure}[h]
    \centering
    \includegraphics[width=0.46\textwidth]{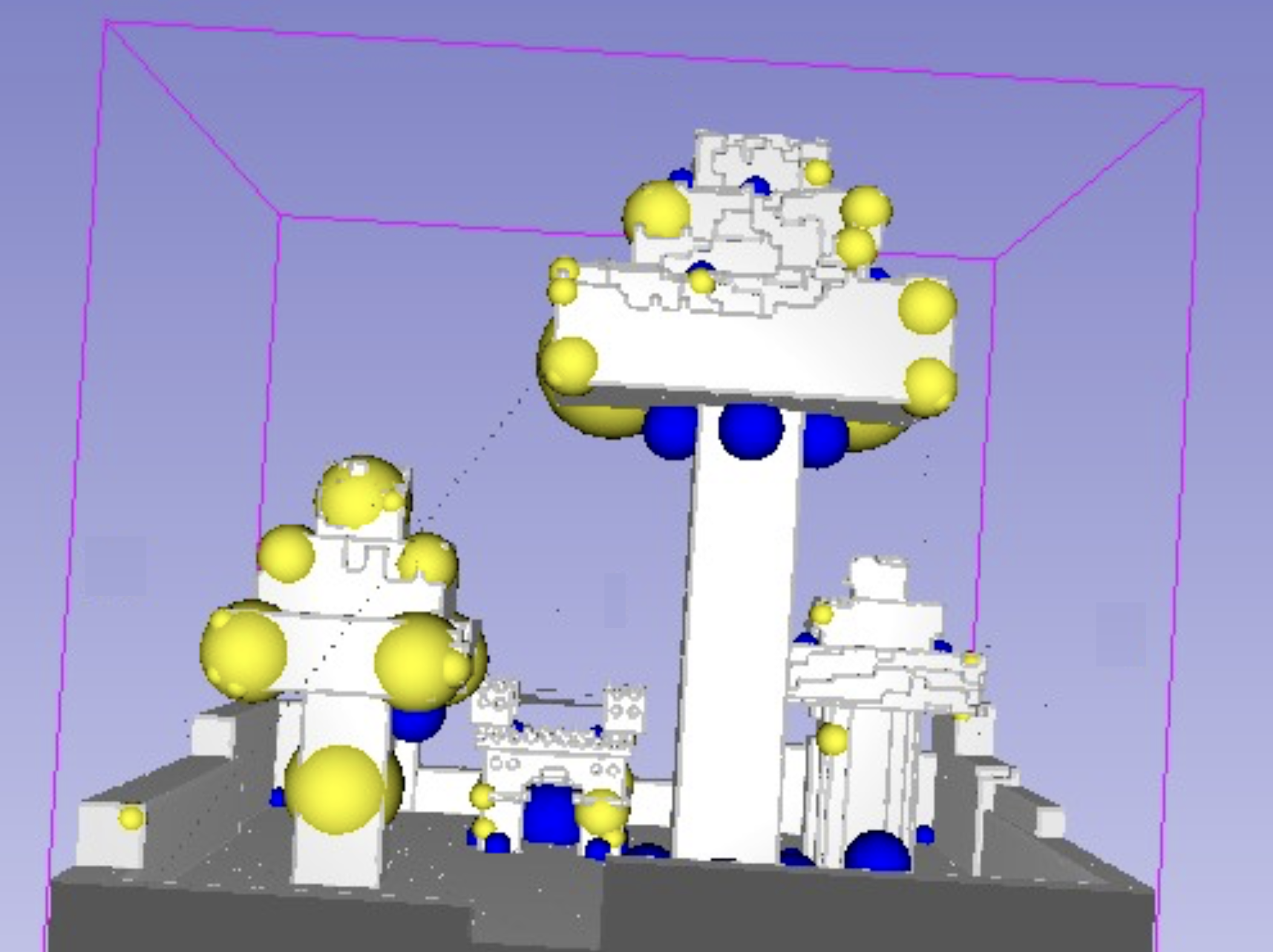} \\
    \includegraphics[width=0.46\textwidth]{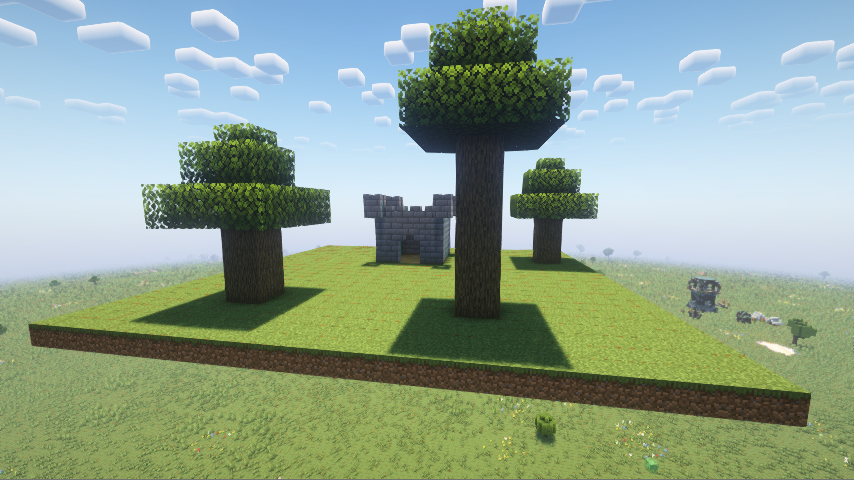}
    \caption{3D SIFT on video game scene (Minecraft).}
    \label{fig:minecraft}
\end{figure}

\bibliography{refs}

\end{document}